\PassOptionsToPackage{table}{xcolor} 
\documentclass[11pt]{article}

\usepackage[preprint]{acl}
\usepackage{marvosym}

\usepackage{times}
\usepackage{latexsym}

\usepackage[T1]{fontenc}

\usepackage[utf8]{inputenc}

\usepackage{microtype}

\usepackage{inconsolata}

\usepackage{graphicx}

\usepackage{graphicx}
\usepackage{booktabs}
\usepackage{multirow}
\usepackage{pifont}
\usepackage{subcaption,soul}
\usepackage{enumitem}
\usepackage{xcolor}

\definecolor{green}{RGB}{69, 189, 155}

\definecolor{dark-green}{rgb}{0,0.5,0}
\definecolor{blue-green}{rgb}{0,0,0.7}

\newcommand{\ourbenchmark}{ChemExplain}

\title{ChemExplain: A Multimodal Benchmark for Explanatory Chemical Reasoning}

\author{
 \textbf{Aotian Chen\textsuperscript{1}},
 \textbf{Frazier N. Baker\textsuperscript{1}},
 \textbf{Reza Averly\textsuperscript{1}},
 \textbf{Shan Zhang\textsuperscript{2}},
 \\
 \textbf{Tianao Chen\textsuperscript{1}},
 \textbf{Daniel Adu-Ampratwum\textsuperscript{1}},
 \textbf{Xia Ning\textsuperscript{1}},
 \textbf{Yuan Xue\textsuperscript{1, \Letter}}
\\
\\
 \textsuperscript{1} The Ohio State University\\
 \textsuperscript{2} Adelaide University
\\
 \small{
   \textsuperscript{\Letter}\textbf{Correspondence:} \href{mailto:xue.643@osu.edu}{xue.643@osu.edu}
 }
}

\begin{document}
\maketitle
\begin{abstract}
Current chemistry benchmarks primarily evaluate outcome prediction or reaction pathway generation, but rarely test whether a model can explain why an observed chemical effect occurs. We introduce \ourbenchmark{}, a benchmark of 1,180 multiple-choice chemistry questions drawn from 383 peer-reviewed papers. Each question pairs a molecular or reaction image with an observed chemical effect and four candidate explanations, each associated with a different structural feature visible in the image. Answering requires the model to perceive the depicted structure, apply chemical knowledge, and identify which feature most directly explains the effect, a capability we term explanatory attribution. The correct feature-mechanism relation is derived from supporting evidence in the source paper rather than invented by the question-generation LLM. We construct the benchmark automatically with a multi-agent framework that applies progressive quality control at three stages and validate it with eight chemistry domain experts. The strongest evaluated model reaches 78.6\% accuracy, leaving considerable room for improvement. On a 600-question expert-reviewed subset, reasoning-trace analysis classifies 71.4\% of the strongest model's errors as attribution failures. Attribution is also the largest error category for the open-source and chemistry-specific models included in our analysis.
\end{abstract}

\section{Introduction}

\begin{figure*}[t]
  \centering
  \includegraphics[width=0.9\textwidth]{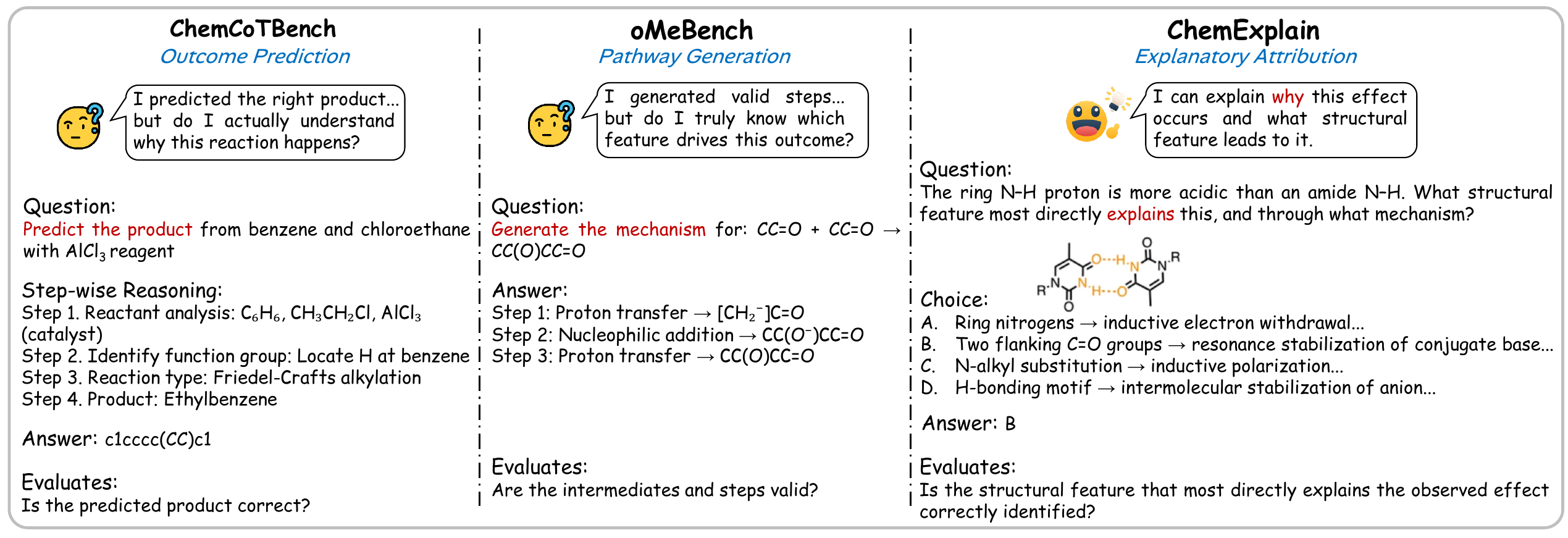}
  \caption{Task-level comparison with representative chemistry benchmarks. ChemCoTBench and oMeBench evaluate reaction outcome prediction and pathway generation, respectively. \ourbenchmark{} instead evaluates whether a model can identify which visible structural feature most directly explains an observed chemical effect.}
  \vspace{-0.2cm}
  \label{fig:comparison}
\end{figure*}

Vision-language models (VLMs) are increasingly used in chemistry for molecular design, retrosynthesis, and reaction optimization, where success depends on reasoning over molecular structures, reaction schemes, and chemical properties. Existing chemistry benchmarks evaluate many of these capabilities~\citep{chembench2024,guo2023ChemLLMBench,huang2024chemeval,zhao2025superchem}, but primarily test reaction outcome prediction or pathway generation~\citep{chemcotbench2024,omebench2024} (Fig.~\ref{fig:comparison}). They rarely ask whether a model can identify which visible structural feature most directly explains an observed chemical effect. A model may predict the correct product while relying on the wrong chemical rationale. This explanatory capability matters because observed effects guide which structural choices to preserve, modify, or discard. Such decisions arise in molecular design, mechanism analysis, and experiment planning, where attributing an effect to the wrong feature leads to modifying the wrong part of a molecule.

Answering such questions requires a multi-step reasoning chain. The model must perceive the relevant structural features in a molecular or reaction image, apply the chemical knowledge needed to relate those features to the observed effect, and determine which feature is primarily responsible in the specific structural context. We refer to this final decision as \emph{explanatory attribution}. Existing benchmarks probe individual parts of this process, including structure recognition~\citep{guo2024molpuzzle}, chemical question answering~\citep{guo2023ChemLLMBench,huang2024chemeval}, and outcome prediction~\citep{chemcotbench2024,omebench2024}, but they do not evaluate the complete process in a single task (Appendix~\ref{sec:related}). They therefore cannot fully distinguish whether a model fails because it misreads the structure, lacks the relevant chemistry, or attributes the effect to the wrong feature.

Studying this capability requires a benchmark grounded in scientific literature, but the necessary evidence is difficult to recover. Chemical structures appear in figures, while the explanation connecting a structure to an observed effect may span captions and body text. Existing extraction tools~\citep{chemdataextractor2016,openchemie2024,mermaid2024,multimodalsearch2025} retrieve structures and associated text but do not determine whether the retrieved text contains a structure-to-effect explanation. We therefore develop a multi-agent framework that isolates individual structures, retrieves and validates explanatory evidence, and applies quality control after each stage. Using this framework, we construct \ourbenchmark{} from about 5,000 chemistry papers while retaining the source paper, figure, structure, and supporting evidence for each selected question.

\ourbenchmark{} contains 1,180 multiple-choice questions drawn from 383 selected peer-reviewed papers. It covers three reasoning types: how a structural feature produces an observed effect (Mechanistic), why a feature was introduced to achieve a target property (Design), and why a structural feature requires particular reaction conditions (Condition) (Appendix Fig.~\ref{fig:distribution}). Each question presents a molecular or reaction image, an observed effect, and four explanations associated with different visible structural features. The model must select the explanation that best accounts for the effect in the depicted structure (Appendix Fig.~\ref{fig:appendix_sample}). The correct feature-mechanism relation is derived from supporting evidence in the source paper rather than invented by an LLM. Benchmark quality is assessed through LLM-as-a-Judge screening and independent evaluation by eight chemistry domain experts.

We evaluate 12 proprietary, open-source, and chemistry-specific VLMs on \ourbenchmark{}. The strongest model reaches 78.6\% accuracy. To locate where models fail, we manually classify the reasoning traces of three representative models on a stratified expert-reviewed subset as perception, knowledge, or attribution failures. Attribution accounts for 71.4\% of the strongest model's errors (Fig.~\ref{fig:error_analysis}) and is the largest category for the open-source and chemistry-specific models, although they make more perception and knowledge errors.

Our contributions are threefold. (1) We introduce \ourbenchmark{}, a literature-grounded benchmark of 1,180 multiple-choice questions for explanatory attribution in chemistry, independently evaluated by eight chemistry domain experts. (2) We present a multi-agent construction framework with progressive quality control and item-level provenance linking each accepted question to its source-paper evidence. (3) We systematically analyze model failures across perception, chemical knowledge, and explanatory attribution, finding that attribution remains the largest error category for all models.

\begin{figure*}[t]
  \centering
  \includegraphics[width=0.95\textwidth]{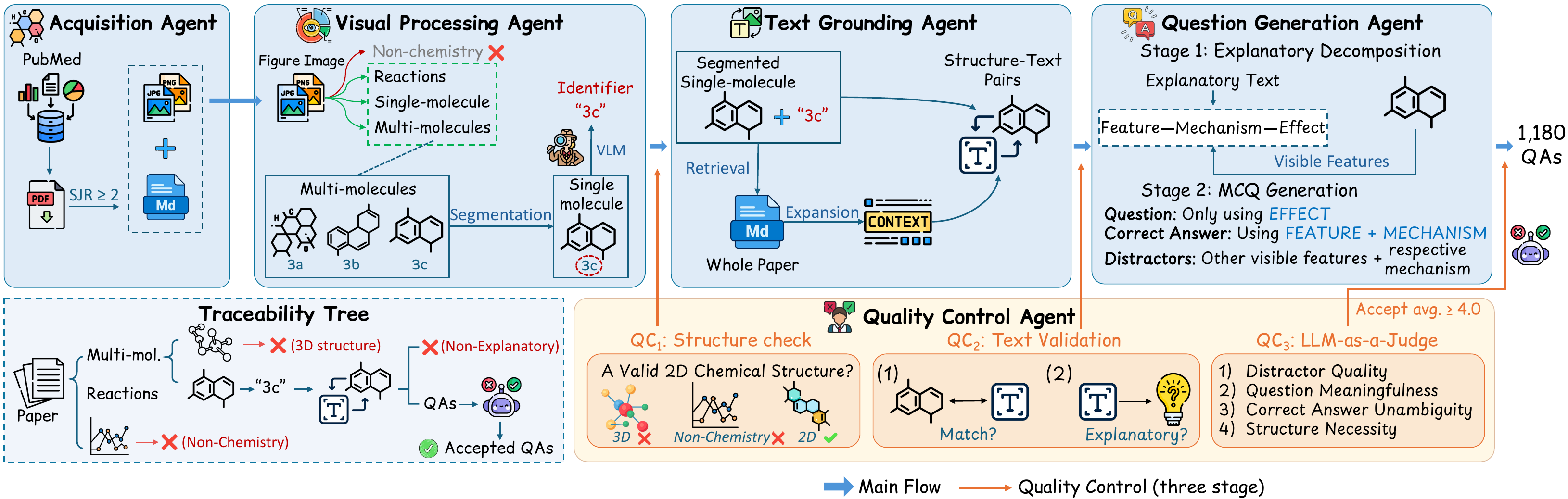}
  \caption{Overview of the multi-agent benchmark construction framework. Four specialist agents perform acquisition, visual processing, text grounding, and question generation, with blue arrows indicating the main data flow. Bottom left: a traceability tree shows how one source paper branches into candidate structures and how invalid candidates are filtered out across stages. Bottom right: a dedicated quality control agent applies three progressive checks between stages: QC$_1$ validates 2D chemical structures, QC$_2$ verifies entity match and explanatory evidence, and QC$_3$ screens generated questions on four quality dimensions.}
  \vspace{-0.4cm}
  \label{fig:method_overview}
\end{figure*}

\section{Related Work}
\label{sec:related_main}

\begin{table}[t]
  \centering
  \small
  \renewcommand{\arraystretch}{1.1}
  \setlength{\tabcolsep}{2pt}
  \renewcommand{\arraystretch}{1.0}
  \resizebox{\linewidth}{!}{%
  \begin{tabular}{l|r|c|c|c|c|c}
    \toprule
    \textbf{Benchmark} & \textbf{Size} & \textbf{Modality} & \textbf{Level} & \textbf{Curation} & \textbf{Source} & \textbf{Attr.} \\
    \midrule
    GPQA~\citep{rein2024gpqa}                   & 448     & Text        & Grad     & Manual & Expert     & \textcolor{red}{\ding{55}} \\
    ChemLLMBench~\citep{guo2023ChemLLMBench}    & 800     & Text        & Mixed    & Semi   & Expert     & \textcolor{red}{\ding{55}} \\
    ChemBench~\citep{chembench2024}             & 2{,}788 & Text        & UG--Grad & Semi   & Expert     & \textcolor{red}{\ding{55}} \\
    ChemCoTBench~\citep{chemcotbench2024}       & 1{,}495 & Text        & Mixed    & Semi   & Expert     & \textcolor{red}{\ding{55}} \\
    oMeBench~\citep{omebench2024}               & 2{,}871 & Text        & Mixed    & Semi   & Expert     & \textcolor{red}{\ding{55}} \\
    SUPERChem~\citep{zhao2025superchem}         & 500     & Multimodal  & Mixed      & Manual & Expert     & \textcolor{red}{\ding{55}} \\
    MolTextQA~\citep{moltextqa2024}             & 500K    & Multimodal  & Mixed    & Auto   & LLM        & \textcolor{red}{\ding{55}} \\
    ChemLitQA~\citep{chemlitqa2024}             & 1{,}411 & Text        & Grad     & Semi   & LLM        & \textcolor{red}{\ding{55}} \\
    \midrule
    \textbf{\ourbenchmark{}} & \textbf{1{,}180} & \textbf{Multimodal} & \textbf{PhD} & \textbf{Auto} & \textbf{Literature} & \textcolor{green!60!black}{\textbf{\ding{51}}} \\
    \bottomrule
  \end{tabular}%
  }
  \caption{\ourbenchmark{} compared with existing chemistry benchmarks. UG = undergraduate, Grad = graduate. \textbf{Source}: origin of the correct answer. \textbf{Attr.}: whether the task requires explanatory attribution.}
  \vspace{-0.5cm}
  \label{tab:benchmark_comparison}
\end{table}

\noindent \textbf{Chemistry Benchmarks.}
Existing chemistry benchmarks mostly test knowledge, prediction, or retrieval, and many are text-only. General scientific benchmarks such as GPQA~\citep{rein2024gpqa} and SciBench~\citep{wang2023scibench} include chemistry questions but do not involve molecular diagrams or reaction schemes. Chemistry-specific benchmarks cover property prediction, structure elucidation, and name conversion~\citep{guo2023ChemLLMBench,chembench2024,huang2024chemeval,guo2024molpuzzle}, while ChemCoTBench~\citep{chemcotbench2024}, oMeBench~\citep{omebench2024}, and SUPERChem~\citep{zhao2025superchem} test reaction outcome prediction and pathway generation. Literature-based QA benchmarks~\citep{moltextqa2024,chemrxivquest2024,scholarchemqa2024,chemlitqa2024} derive questions from papers but mainly test factual retrieval. These benchmarks rarely ask which visible structural feature explains an observed effect. \ourbenchmark{} targets this capability (Tab.~\ref{tab:benchmark_comparison}).

\noindent \textbf{Benchmark Construction.}
Benchmarks also differ in where the correct answer comes from. Some literature-based QA benchmarks use LLMs to generate reference answers, introducing a risk of unsupported explanations when those answers are not independently verified~\citep{chemlitqa2024}. Agentic pipelines such as AutoConverter~\citep{zhang2025automated} and MicroVQA~\citep{burgess2025microvqa} instead convert human-written question--answer pairs into multiple-choice questions. Deriving the answer from a paper requires locating the text that explains the effect, which extraction tools~\citep{chemdataextractor2016,openchemie2024,mermaid2024,multimodalsearch2025} do not separate from procedural or descriptive text. \ourbenchmark{} locates and validates this explanatory evidence and derives each correct answer from it (Appendix~\ref{sec:related}).

\section{Method}
\label{sec:method}

Our goal is to construct benchmark questions from chemistry papers that ask which visible structural feature most directly explains an observed chemical effect. Each question pairs a molecular or reaction image with an effect and a set of candidate explanations, and remains traceable to the supporting evidence in the source literature.

Figure~\ref{fig:method_overview} summarizes the multi-agent construction framework. Four specialist agents perform paper acquisition, visual processing, text grounding, and question generation. The acquisition agent converts PDFs into structured text and figure images with MinerU~\citep{mineru2024}. The visual processing agent identifies chemistry figures, segments them into molecular or reaction instances, and recovers paper-specific identifiers such as ``compound 4b.'' The text grounding agent uses each identifier to retrieve evidence and determine whether it contains a structure-to-effect explanation. The question generation agent converts validated structure-evidence pairs into multiple-choice items while preventing answer leakage.

A quality control agent checks the output after each core stage. QC$_1$ removes invalid or non-2D chemical structures. QC$_2$ verifies entity match and explanatory value. QC$_3$ screens each generated item on four quality dimensions (Section~\ref{sec:quality_validation}). The provenance tree records each candidate and its quality-control decisions, linking every accepted question to its source paper, figure, structure, and supporting evidence. Sections~\ref{sec:visual}--\ref{sec:synthesis} describe the three stages in detail, with additional implementation information in Appendix~\ref{sec:impl_details}.

\subsection{Visual Processing}
\label{sec:visual}

Chemistry figures often contain multiple molecular diagrams or reaction schemes, whereas each benchmark item must be linked to one molecule or reaction. The visual processing stage therefore isolates individual structures and recovers the identifiers that connect them to the article text.

A ResNet-50 classifier~\citep{he2016resnet} separates chemistry figures from spectra, plots, and experimental images. DECIMER Segmentation~\citep{rajan2021decimer} isolates structures in multi-molecule figures, while RxnScribe~\citep{qian2023rxnscribe} decomposes reaction schemes. We retain the segmented structures as images because SMILES conversion disproportionately removes complex structures, including many cases with detailed mechanistic discussion (Section~\ref{sec:ablation}). Existing co-reference tools such as MolCoref~\citep{openchemie2024} recover identifiers for only a minority of our figures. We instead highlight each target structure in the original figure and use GPT-4o-mini to recover its identifier. Section~\ref{sec:ablation} compares these alternatives.

\subsection{Text Grounding}
\label{sec:text}

From the source literature corpus, we observe that most textual mentions of a molecule are procedural or descriptive rather than explanatory. Thus, we require evidence that connects a visible structural feature to an observed effect. GPT-4o-mini searches the full paper using the structure identifier, image, caption, and in-text figure references. Because the effect and mechanism often appear in adjacent sentences, each retrieved sentence is expanded with semantically related sentences from the same paragraph. Section~\ref{sec:ablation} quantifies the contribution of this context expansion.

QC$_2$ evaluates both \emph{entity match} and \emph{explanatory value}. Entity match indicates whether the passage refers to the target molecule or reaction. Explanatory passages state how a structural feature produces the observed effect. Partially explanatory passages identify the feature but omit the mechanism, while non-explanatory passages are procedural or descriptive. We discard entity mismatches and non-explanatory passages. Partial explanations are retained only as context, and a structure advances to question generation only if its evidence block contains an explanatory sentence.

\subsection{Question Generation}
\label{sec:synthesis}

The question generation agent converts each validated structure-evidence pair into a multiple-choice question. The source paper determines the correct feature-mechanism relation, while GPT-5.1 reformulates this evidence into the question and answer options. The model is not allowed to introduce a correct-answer mechanism that is absent from the supporting evidence. The distractors use other genuine features visible in the same structure.

\noindent \textbf{Two-stage generation.}
GPT-5.1 first decomposes the explanatory evidence into three fields: (1)~Effect, the observed outcome without naming its structural cause (e.g., ``resists oxidative degradation''); (2)~Feature, the feature identified by the source paper as primarily responsible; and (3)~Mechanism, how that feature produces the effect. The question stem is generated from Effect alone, so Feature and Mechanism cannot leak into the wording of the stem. The correct option restates the Feature and Mechanism fields. Each distractor pairs another visible structural feature with a chemically plausible mechanism for the same effect.

\noindent \textbf{Anti-shortcut constraints.}
We apply two constraints to reduce textual shortcuts without reading the image content. \emph{Same-effect competition} requires every option to explain the same observed effect, reducing the possibility of identifying the correct option through simple textual matching. \emph{Matched confidence} requires comparable tone and assertiveness across options and excludes hedging terms that could reveal distractors. Answer positions are randomized. Appendix~\ref{sec:prompts} provides the complete prompts.

\section{\ourbenchmark{} Construction and Validation}
\label{sec:validation}

\subsection{Source Corpus and Construction Yield}
\label{sec:construction}

\begin{figure}[t]
  \centering
  \includegraphics[width=\columnwidth]{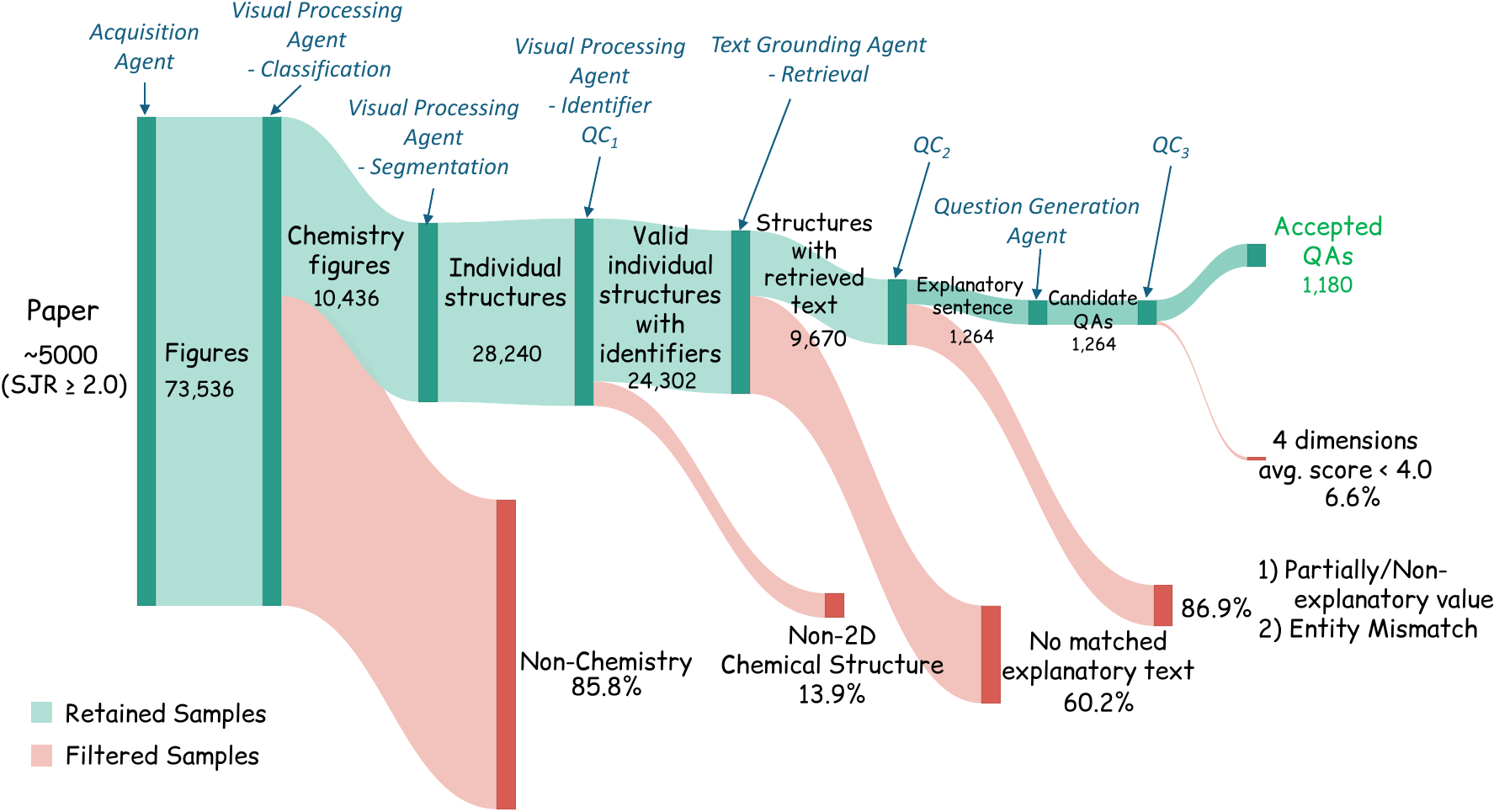}
  \caption{Construction yield and attrition from $\sim$5,000 source papers to 1,180 accepted questions. Green denotes retained samples, red denotes filtered samples annotated with rejection cause and loss percentage, and stage labels indicate the responsible agent.}
  \vspace{-0.2cm}
  \label{fig:pipeline_yield}
\end{figure}

To instantiate the framework at scale, we draw source papers from the PubMed Central Open Access Subset (PMC OA), which provides articles under licenses that permit text mining and derivative use. We query PMC for papers in organic synthesis, materials chemistry, catalysis, and medicinal chemistry, yielding approximately 24,000 candidate papers. We further restrict the source corpus to journals with SCImago Journal Rank (SJR) $\geq 2.0$~\citep{scimago2024}, leaving approximately 5,000 papers in the final source corpus. Fig.~\ref{fig:pipeline_yield} summarizes the construction yield at each stage.

From this filtered corpus, the acquisition agent extracts 73,536 figures. After figure classification, segmentation, identifier extraction, and structure validation, the visual processing agent yields 24,302 valid molecular or reaction instances spanning 2,906 papers. The text grounding agent retrieves and validates explanatory evidence for these instances: 9,670 (39.8\%) yield retrieved text, and 1,264 of these (13.1\%) pass both entity-match and explanatory-value validation, which qualifies the instance for question generation. The question generation agent converts these 1,264 candidates into questions, of which 1,180 (93.4\%) pass the quality control agent's LLM-as-a-Judge screening (Section~\ref{sec:quality_validation}) and are accepted into the final benchmark. Section~\ref{sec:pipeline_analysis} analyzes where these losses occur.

\subsection{Dataset Statistics}
\label{sec:statistics}

The final benchmark contains 1{,}180 accepted questions drawn from 383 peer-reviewed papers (Appendix Fig.~\ref{fig:distribution}). The three reasoning categories are: \emph{Mechanistic}, which asks how a structural feature leads to an observed effect (55.9\%); \emph{Design}, which asks why a feature was introduced to achieve a target property (39.7\%); and \emph{Condition}, which asks why a structural feature requires specific reaction conditions (4.3\%). In terms of entry type, the benchmark contains 338 molecule questions (28.6\%) and 842 reaction questions (71.4\%). These proportions follow the source literature, in which explicit structure-to-condition explanations are uncommon. Each paper contributes 3.08 questions on average (median 3, maximum 8). Appendix~\ref{sec:question_examples} shows examples of each category and entry type.

The benchmark is comparable in size to expert-curated reasoning benchmarks such as GPQA (448) and SUPERChem (500), and its automated construction supports extension to new source papers and subdomains without manual question writing.

\begin{table}[t]
  \centering
  \small
  \renewcommand{\arraystretch}{0.8}
  \resizebox{\columnwidth}{!}{%
  \setlength{\tabcolsep}{10pt}
  \begin{tabular}{l|c|ccc|c}
    \toprule
    \multirow{2}{*}[-0.3em]{\textbf{Dimension}}
      & \multirow{2}{*}[-0.3em]{$\bar{H}$}
      & \multicolumn{3}{c|}{AC$_2$}
      & \multirow{2}{*}[-0.3em]{MAE$\downarrow$} \\
    \cmidrule(lr){3-5}
      & & H--H & H--L & SE & \\
    \midrule
    Distractor Quality & 4.50 & 0.909 & 0.948 & 0.009 & 0.497 \\
    Meaningfulness & 4.42 & 0.898 & 0.936 & 0.009 & 0.505 \\
    Answer Unambiguity & 4.43 & 0.899 & 0.929 & 0.011 & 0.407 \\
    Struct.\ Necessity & 4.75 & 0.946 & 0.861 & 0.022 & 0.785 \\
    \midrule
    Avg.\ (Score) & 4.53 & 0.913 & 0.919 & 0.013 & 0.549 \\
    \bottomrule
  \end{tabular}
  }
  \caption{LLM-as-a-Judge reliability ($n{=}600$ stratified sample from accepted questions, 8 human experts). $\bar{H}$: human expert mean; AC$_2$: Gwet's AC$_2$; H--H: human inter-rater; H--L: human--LLM; SE: analytic standard error of the H--L estimate; MAE: mean absolute error.}
  \vspace{-0.4cm}
  \label{tab:agreement}
\end{table}

\subsection{Quality Validation}
\label{sec:quality_validation}

We evaluate benchmark quality through a two-phase protocol that combines full-corpus LLM screening with expert assessment of a stratified subset. Human experts and the LLM judge independently rate the same 600 questions using the same four quality dimensions. Their agreement measures how closely the automated screening reproduces expert quality judgments.

\noindent \textbf{LLM-as-a-Judge Screening.}
We use GPT-5.2 as the judge model with explanation-first prompting~\citep{zheng2023mtbench,pan2024judgebench}, so that the model first analyzes each dimension before assigning a numeric score. The judge evaluates all 1,264 candidate questions on a 1--5 Likert scale along four dimensions: \emph{Distractor Quality}, whether incorrect options remain plausible without careful analysis of the depicted structure; \emph{Question Meaningfulness}, whether the question probes an explanatory relationship rather than superficial recall; \emph{Correct Answer Unambiguity}, whether the labeled answer is the best-supported explanation and no competing option is equally defensible; and \emph{Structure Necessity}, whether solving the question requires the depicted structure rather than the question text alone. To reduce answer-order effects~\citep{shi2025positionbias}, each question is scored three times with shuffled option order, and the scores are averaged. Of the 1,264 candidates, 1,180 (93.4\%) achieve an average score of at least 4.0 and are accepted. The remaining 84 are rejected (Appendix~\ref{sec:failure_table} analyzes the rejections and Appendix~\ref{sec:construction_failures} shows examples).

\noindent \textbf{Human Expert Evaluation.}
Eight chemistry experts, including five faculty members and three PhD or postdoctoral researchers, independently evaluate a stratified sample of 600 accepted questions. The sample contains 300 molecule questions and 300 reaction questions, includes all 51 Condition questions together with 288 Mechanistic and 261 Design questions, and uses different source papers whenever possible. All eight experts rate every question on every dimension, yielding 4,800 expert-question ratings per dimension (Tab.~\ref{tab:agreement}). Each evaluator first answers the question without access to the source-paper text and then reviews the supporting evidence before assigning quality scores on the same 1--5 scale (Appendix~\ref{sec:eval_platform}). In this blind phase, at least five of the eight experts select the labeled answer for 94.0\% of questions (Appendix~\ref{sec:human_expert}).

\noindent \textbf{Agreement and Validation Results.}
We compare judge and expert scores to determine whether the judge and the experts apply the quality rubric consistently. The mean judge scores on the expert subset are close to those on the full accepted benchmark after Holm--Bonferroni correction, with absolute differences of 0.003--0.125 points across dimensions (Appendix~\ref{sec:sample_balance}). We report Gwet's AC$_2$ with quadratic weights~\citep{gwet2014handbook} because ratings cluster near the top of the scale, where weighted $\kappa$ and Krippendorff's $\alpha$ suffer from the prevalence paradox, together with MAE and confidence intervals (\citealp{klie2024analyzing}; Appendix~\ref{sec:agreement_metric}).

Tab.~\ref{tab:agreement} summarizes the results. The experts assign a mean quality score of 4.53 across dimensions, and every question also passes the 4.0 acceptance threshold under expert scores. Human-human agreement is high, with an average AC$_2$ of 0.913, indicating that the rubric is applied consistently across experts. Human-LLM agreement is similar, with an average AC$_2$ of 0.919 and all four dimensions above 0.86. The judge also agrees with the experts on rejected candidates: on 60 rejected questions that the experts rate without knowing the judge's decision, both assign a mean score of 3.68 (Appendix Tab.~\ref{tab:failure}). These results support using the judge as a scalable quality-control stage for all 1,264 candidates. Claude-Opus-4-7 and Gemini-3.1-Pro yield comparable or higher agreement on the same subset (Appendix~\ref{sec:judge_robustness}), showing that the result is not specific to the GPT judge.

\section{Experiments and Analysis}
\label{sec:experiments}

\subsection{Main Results}

We evaluate \ourbenchmark{} on a diverse set of vision-language models, including closed-source systems (Gemini-3.1-Pro~\citep{gemini2026}, GPT-5.4, GPT-5, GPT-4o~\citep{openai2025gpt5,openai2023gpt4}, Claude Sonnet~4.6~\citep{anthropic2025claude}), open-source models (Qwen3-VL~\citep{bai2025qwen3vl}, Qwen2.5-VL~\citep{bai2025qwen25vl}, Phi-4-reasoning-vision~\citep{aneja2026phi4reasoning}, InternVL3.5~\citep{zhu2025internvl3}, Janus-Pro~\citep{chen2025januspro}), and one chemistry-specific model (ChemVLM~\citep{li2025chemvlm}). All models are evaluated in the zero-shot setting with majority-vote accuracy over three independent trials, following common practice in multimodal benchmark evaluation~\citep{yue2024mmmu}. A question counts as incorrect when the three trials yield three different answers or when an output cannot be parsed into an option. We use reasoning-first prompting, instructing each model to analyze structural features and their mechanistic roles before producing a final answer. As a human reference, the eight chemistry experts answer the same 600-question subset used in Section~\ref{sec:quality_validation}, following prior benchmarks with expert baselines on a subset~\citep{chembench2024}. These results serve as a reference for answerability and difficulty rather than a direct human-model comparison.

\begin{table}[t]
  \centering
  \footnotesize   
  \renewcommand{\arraystretch}{1.0}
  \setlength{\tabcolsep}{1.6mm}
  \resizebox{\linewidth}{!}{%
    \begin{tabular}{l|c|c|ccc|cc}
      \hline
      \multirow{2}{*}{\textbf{Model}} & \multirow{2}{*}{\textbf{Size}} & \multirow{2}{*}{\textbf{Avg.}} & \multicolumn{3}{c|}{\textbf{Reasoning Type}} & \multicolumn{2}{c}{\textbf{Entry Type}} \\ \cline{4-8}
       & & & \textit{Mech.} & \textit{Design} & \textit{Cond.} & \textit{Mol.} & \textit{Rxn.} \\ \hline
      \textit{Random baseline} & -- & 25.0 & 25.0 & 25.0 & 25.0 & 25.0 & 25.0 \\ \hline
      \rowcolor{green!15}\multicolumn{8}{l}{\textit{Proprietary VLMs}} \\ \hline
      GPT-5.4 & -- & \textbf{78.6} & 75.3 & \textbf{84.2} & 68.6 & \textbf{81.1} & \textbf{77.6} \\
      GPT-5 & -- & 77.4 & 72.9 & 83.6 & \textbf{78.4} & 80.2 & 76.2 \\
      GPT-4o & -- & 68.1 & 65.6 & 72.5 & 60.8 & 68.6 & 67.9 \\
      Claude Sonnet 4.6 & -- & 78.2 & \textbf{76.6} & 81.1 & 70.6  & 79.9 & 77.5 \\
      Gemini-3.1-Pro & -- & 78.0 & 76.1 & 80.6 & \textbf{78.4} & 80.2 & 77.1 \\ \hline
      \rowcolor{green!15}\multicolumn{8}{l}{\textit{Open-Source VLMs}} \\ \hline
      Qwen2.5-VL & 7B & 42.6 & 38.5 & 48.8 & 39.2 & 41.4 & 43.1 \\
      Qwen3-VL & 8B & 52.5 & 47.7 & 59.5 & 49.0 & 51.8 & 52.7 \\
      Qwen3-VL & 30B & 58.8 & 54.7 & 65.0 & 54.9 & 62.4 & 57.4 \\
      Phi-4-reasoning-vision & 15B & 51.3 & 50.6 & 52.2 & 51.0 & 45.9 & 53.4 \\
      InternVL3.5 & 8B & 48.0 & 45.6 & 52.0 & 43.1 & 47.9 & 48.1 \\
      Janus-Pro & 7B & 23.2 & 20.0 & 27.5 & 25.5 & 26.3 & 22.0 \\ \hline
      \rowcolor{green!15}\multicolumn{8}{l}{\textit{Chemistry-Specific VLMs}} \\ \hline
      ChemVLM & 8B & 34.0 & 33.2 & 35.2 & 33.3 & 28.4 & 36.2 \\  \hline
    \end{tabular}%
  }
  \caption{Performance on \ourbenchmark{} (majority-vote accuracy, \%, 3 trials). Mech.\ = Mechanistic, Cond.\ = Condition, Mol.\ = molecule, Rxn.\ = reaction. Best per column in \textbf{bold}. All results are on the full 1{,}180-question benchmark. Human expert performance is reported separately in Appendix Tab.~\ref{tab:human_expert}.}
  \vspace{-0.4cm}
  \label{tab:main_results}
\end{table}

\noindent \textbf{Overall performance.} 
Tab.~\ref{tab:main_results} presents the main results. On the 600-question subset, the experts average 73.3\% and range from 68.3\% to 80.0\% (Appendix Tab.~\ref{tab:human_expert}), confirming that the questions are answerable for trained chemists but remain challenging. Differences among experts partly reflect domain specialization across materials, polymer, inorganic, and other areas of chemistry: on questions within his own domain, one expert reaches 94.1\% (Appendix~\ref{sec:fine_grained}). On the full benchmark, the top three models cluster near 78\% (GPT-5.4: 78.6\%, Claude Sonnet 4.6: 78.2\%, Gemini-3.1-Pro: 78.0\%), with no statistically significant pairwise differences detected (Appendix~\ref{sec:paper_level}). The strongest open-source model, Qwen3-VL-30B, reaches 58.8\%, nearly 20 points lower. Scaling Qwen3-VL from 8B to 30B improves accuracy from 52.5\% to 58.8\%, but leaves a substantial gap. Appendix~\ref{sec:reasoning_type} reports accuracy by reasoning type.

\noindent \textbf{Domain-specific VLM remains weak on explanatory attribution.}
ChemVLM reaches 34.0\%, below every general-purpose open-source model except Janus-Pro. Its errors span perception, knowledge, and attribution (Section~\ref{sec:error_analysis}), so the performance gap is not limited to the final attribution step. Model size does not fully explain the result: the size-matched InternVL3.5-8B and Qwen3-VL-8B reach 48.0\% and 52.5\%, respectively. ChemVLM is trained mainly for structure recognition and exam-style QA, whereas \ourbenchmark{} requires visually grounded reasoning from a structural feature to an observed effect.


\subsection{Visual Necessity}
\label{sec:visual_necessity}

Multimodal multiple-choice benchmarks can contain textual shortcuts that allow questions to be answered without the image~\citep{yue2025mmmu}. We test the visual necessity targeted by the anti-shortcut design (Section~\ref{sec:synthesis}) on the 600-question expert subset (Section~\ref{sec:quality_validation}). We remove only the structure image and keep the question and all four options unchanged, which preserves all lexical, stylistic, and elimination cues. As shown in Tab.~\ref{tab:text_only}, accuracy falls by 53.5--54.0 points across the three models, from 74.0--78.5\% with the image to 20.5--24.5\% without it. Without the image, all three models perform below the 25\% random baseline, suggesting that the question and options provide little exploitable information for these models. The substantial performance drop confirms that their success depends largely on the depicted structure.

\begin{table}[t]
\centering
\renewcommand{\arraystretch}{1.0}
\small
\resizebox{0.75\columnwidth}{!}{%
\begin{tabular}{@{}l|c|c|c@{}}
\hline
\textbf{Model} & With image & Text-only & $\Delta$ \\
\hline
GPT-5 & 77.0 & 23.5 & $-$53.5 \\
Claude Sonnet 4.6 & 74.0 & 20.5 & $-$53.5 \\
Gemini-3.1-Pro & 78.5 & 24.5 & $-$54.0 \\
\hline
\end{tabular}%
}
\caption{Text-only control on the 600-question expert subset, removing only the structure image.}
\vspace{-0.4cm}
\label{tab:text_only}
\end{table}

\subsection{VLM Error Analysis}
\label{sec:error_analysis}

To understand where in the reasoning chain (perception $\rightarrow$ knowledge $\rightarrow$ attribution) models fail, we adapt standard multimodal error taxonomies~\citep{yue2024mmmu,yue2025mmmu} into a domain-specific taxonomy. A chemistry faculty member manually analyzes responses from three representative models---GPT-5.4 (best closed-source), Qwen3-VL-8B (representative open-source), and ChemVLM (chemistry-specific)---on the same 600-question stratified subset (Section~\ref{sec:quality_validation}), categorizing each incorrect response by its reasoning trace as a \emph{perception error} (the model misidentifies or overlooks visible structural features), a \emph{knowledge gap} (it perceives the features but lacks the chemical knowledge linking them to the effect), or an \emph{attribution error} (it perceives the relevant features and has the necessary knowledge but misjudges which feature is primarily responsible in this structural context).

Three chemistry experts independently reclassified all 850 incorrect responses. They assigned the same error type in 96.5\% of cases, with Gwet's AC$_2$=0.984 (95\% CI $[0.966,1.000]$). This high agreement supports the consistency of the perception, knowledge, and attribution taxonomy.

\begin{figure}[t]
  \centering
  \includegraphics[width=0.8\columnwidth]{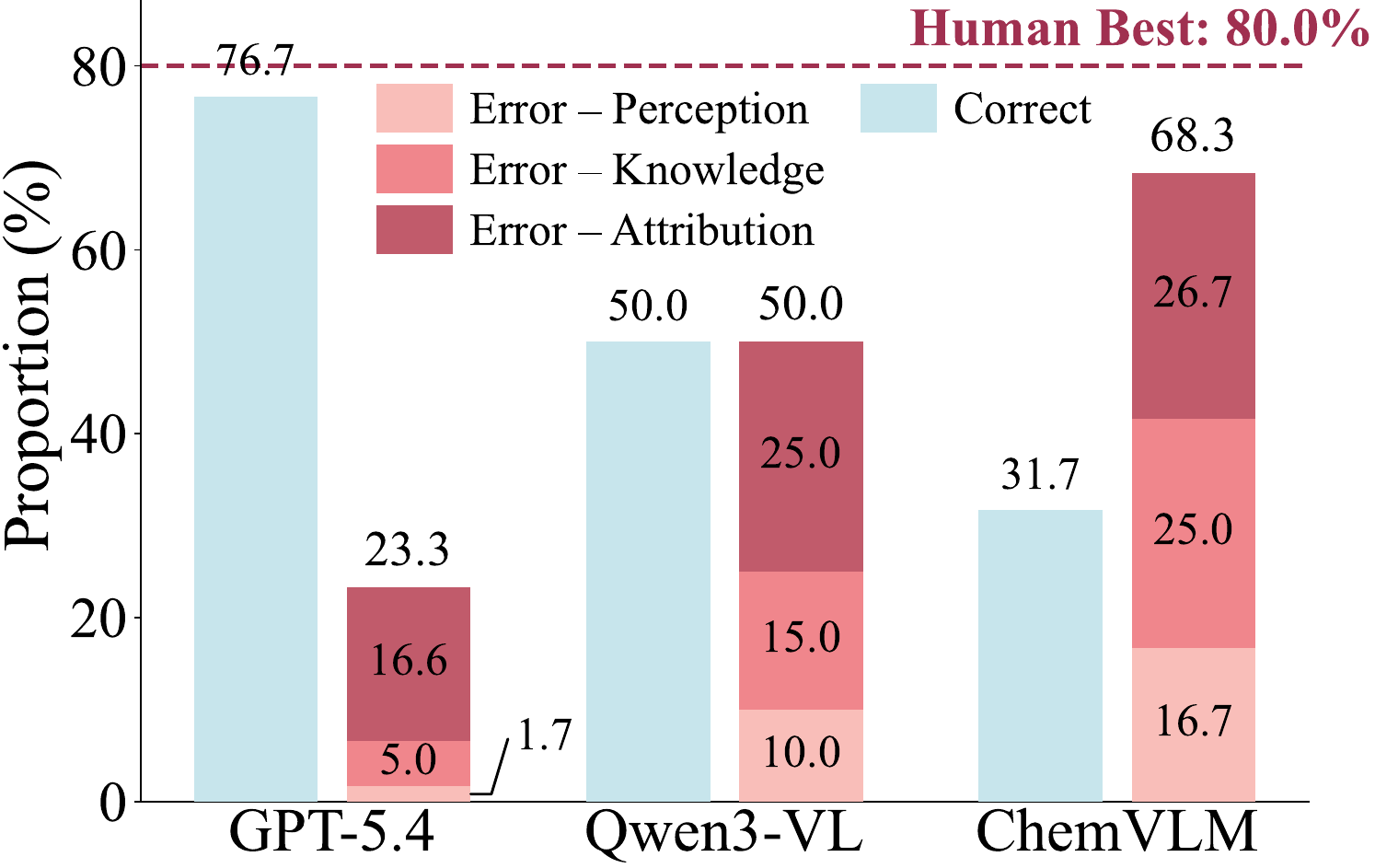}
  \caption{Accuracy and error type breakdown for three representative models on the 600-question evaluation subset. Dashed line: best human expert accuracy.}
  \vspace{-0.4cm}
  \label{fig:error_analysis}
\end{figure}

\noindent \textbf{Error distribution.}
Across the three models, attribution accounts for 48\% of all incorrect responses, followed by knowledge gaps (32\%) and perception errors (20\%) (Fig.~\ref{fig:error_analysis}). The error profiles differ with model performance. GPT-5.4 makes only 10 perception errors on the subset, while 100 of its 140 errors (71.4\%; 95\% Wilson CI {$[63.5, 78.3]$}) are attribution errors, a proportion that remains 72.0\% on questions with strong expert consensus in the blind phase (Appendix~\ref{sec:human_expert}). In these attribution-classified failures, the reasoning traces correctly identify relevant structural features and apply plausible chemistry, but select the wrong feature as the primary explanation. Providing the structural features and the relevant chemistry corrects most perception errors and knowledge gaps, but attribution errors largely persist (Appendix~\ref{sec:intervention}). A chemistry expert answering without the source paper resolves 60 of GPT-5.4's 100 attribution errors, and most of those he misses fall outside his domain (Appendix~\ref{sec:fine_grained}). Qwen3-VL-8B shows a more balanced error profile, with attribution accounting for 50\% of its failures. ChemVLM makes 100 perception errors and 150 knowledge errors, showing that its low performance cannot be attributed to the final attribution step alone. Appendix~\ref{sec:error_cases} presents representative examples of all three error types.


\subsection{Agentic Construction Pipeline Analysis}
\label{sec:pipeline_analysis}
\label{sec:ablation}

We analyze the main construction choices on 100 randomly sampled source papers (Tab.~\ref{tab:ablation}) and examine attrition across the full corpus (Fig.~\ref{fig:pipeline_yield}). Repeating the subset analysis with Claude Sonnet 4.6 supports the same design choices (Appendix~\ref{sec:ablation_claude}), and Appendix~\ref{sec:pipeline_limitations} shows that current VLMs struggle to distinguish structural analogs and to identify explanatory evidence in text.

\begin{table}[t]
  \centering
  \small
  \renewcommand{\arraystretch}{1.0}
  \setlength{\tabcolsep}{4pt}
  \resizebox{\columnwidth}{!}{%
  \begin{tabular}{l|c|c|c}
    \toprule
    \textbf{Configuration} & \textbf{Structures} & \textbf{Ent.\ Match} & \textbf{Expl.} \\
    \midrule
    Full pipeline (ours) & 489 & 76\% & 43 \\
    \midrule
    \quad w/o image-first (SMILES mode) & 154 & --- & 2 \\
    \quad w/o VLM identifier & 489 & 50\% & 34 \\
    \quad w/ MolCoref identifier & 489 & 35\% & 31 \\
    \quad w/o context expansion & 489 & 76\% & 26 \\
    \bottomrule
  \end{tabular}%
  }
  \caption{Construction pipeline ablation on 100 randomly sampled source papers. Each row below the full pipeline removes or replaces one component, and the columns report pipeline outputs. Structures: individual molecular or reaction images retained after segmentation. Ent. Match: fraction of retrieved text referring to the target molecule or reaction. Expl.: explanatory sentences.}
  \vspace{-0.4cm}
  \label{tab:ablation}
\end{table}

\noindent \textbf{Image-first representation preserves structurally complex molecules.}
A central design choice is whether to retain segmented structure images or first convert them into SMILES strings. The SMILES-first alternative, implemented with MolScribe~\citep{qian2023molscribe} and DECIMER~\citep{rajan2021decimer} plus round-trip quality control (Appendix~\ref{sec:smiles_pipeline}), discards 68.5\% of the retained structures and leaves only 2 of the 43 explanatory sentences (Tab.~\ref{tab:ablation}). Extraction failures are particularly common for large rings, densely substituted structures, and nonstandard drawing conventions, including structures mechanistically discussed in the literature. Retaining images therefore avoids a strong selection bias and preserves structural diversity.

\noindent \textbf{Identifier-based retrieval enables precise text grounding.}
Removing VLM-based identifier extraction lowers entity match from 76\% to 50\%, and substituting MolCoref~\citep{openchemie2024} lowers it further to 35\% (Tab.~\ref{tab:ablation}). MolCoref often assigns no identifier or the identifier of a neighboring compound, which misleads downstream retrieval.

\noindent \textbf{Context expansion recovers explanatory content across sentence boundaries.}
Without context expansion, only 26 explanatory sentences are identified in the 100 sampled papers, whereas expansion raises this number to 43 (a 65\% gain). Even so, the availability of explanatory text remains the primary bottleneck: in the full corpus, only 13.1\% of structures with retrieved text contain at least one explanatory sentence. To verify that this attrition reflects a scarcity of explanatory text rather than a retrieval failure, a chemistry expert audited 244 candidate structures from 50 random papers and found the same rates as the full corpus: 39.8\% had retrieved text, and 13.4\% of these contained an explanatory sentence (13.1\% in the full corpus). The discarded cases lack a structure-to-effect explanation rather than being mis-retrieved, and most are partially explanatory, naming a feature and its effect without the mechanism (Appendix~\ref{sec:pipeline_limitations}).


\section{Discussion}
\label{sec:discussion}

\noindent \textbf{What \ourbenchmark{} measures.}
\ourbenchmark{} evaluates explanatory attribution: identifying which visible structural feature most directly explains an observed chemical effect. Each question isolates this decision, because all four options explain the same effect through features visible in the structure (Section~\ref{sec:synthesis}), so choosing among them requires judging which feature is primarily responsible. Two results show that the benchmark measures this decision reliably. The questions require the depicted structure, as removing the image lowers accuracy to 20.5--24.5\% (Section~\ref{sec:visual_necessity}). The labeled answers reflect expert consensus, as at least five of the eight experts select them for 94.0\% of questions without seeing the source paper (Section~\ref{sec:quality_validation}).

\noindent \textbf{Implications for model development.}
Attribution errors decrease least as models become stronger: from ChemVLM to GPT-5.4 they fall by 38\%, compared with 80--90\% for perception and knowledge errors, and they persist when the structure and the relevant chemistry are supplied (Section~\ref{sec:error_analysis}). These findings suggest that training models to compare competing explanations within a specific structural context may improve attribution. The Effect, Feature, and Mechanism fields in ChemExplain provide a natural source of supervision for this purpose. Evidence-Grounded Attribution, which links each option to evidence in the depicted structure, already improves three open-source models by 3.1--5.3 points, with the largest gain for ChemVLM (Appendix~\ref{sec:ega}). Chemistry-specific VLMs thus need training data pairing structures with explanations of their effects, beyond structure recognition and exam-style QA.


\section{Conclusion}
\label{sec:conclusion}


We introduced \ourbenchmark{}, a literature-grounded benchmark of 1,180 questions for explanatory attribution in chemistry, constructed automatically from peer-reviewed papers and assessed by eight chemistry experts. The strongest evaluated model reaches 78.6\% accuracy, and reasoning-trace analysis classifies 71.4\% of its errors as attribution failures. These results identify explanatory attribution as an unresolved problem in multimodal chemical reasoning and establish \ourbenchmark{} as a practical way to measure progress. Appendix~\ref{sec:limitations} discusses design choices and future directions.

\section*{Acknowledgments}
This work was supported in part by the National Science Foundation (NSF) under Grant No. 2435819 and the National Library of Medicine (NLM) of the National Institutes of Health (NIH) under Award No. R01LM014385.

\section*{Limitations}

\ourbenchmark{} evaluates explanatory attribution in a multiple-choice format, which enables automatic and reproducible scoring. Free-form explanation generation is a complementary setting, and the source evidence retained for each question could serve as its reference. The benchmark focuses on 2D structure drawings, and extending it to other representations such as spectra, 3D conformations, or crystal structures is a direction for future versions.

\section*{Ethical Considerations}
\label{sec:ethics}

All source papers are drawn from the PMC Open Access Subset, which distributes articles under various Creative Commons licenses (CC-BY, CC-BY-NC, CC-BY-SA, and others) that permit text mining and derivative works. Our benchmark construction constitutes a derivative work that extracts and transforms content from these articles. Each benchmark item retains full provenance to its source paper, enabling verification and compliance with individual article license terms. We release \ourbenchmark{} under CC-BY-NC-SA 4.0 for research purposes. Users intending commercial applications should consult the license of each source article. To support transparency and reproducibility, we will also release the construction code and the full prompt templates upon publication.


\bibliography{main}

\appendix
\twocolumn[
\begin{center}
  {\Large\bfseries ChemExplain: A Multimodal Benchmark for Explanatory Chemical Reasoning}\\
  \vspace{0.5em}
  {\Large Supplementary Materials}
  \vspace{3em}
\end{center}
]

\noindent\textbf{Table of Contents}
\begin{itemize}[leftmargin=1.5em, labelwidth=1.5em, labelsep=0em,
                align=left, itemsep=2pt]
  \item[\textbf{A.}] Extended Related Work
  \item[\textbf{B.}] Representative Benchmark Example
  \item[\textbf{C.}] Construction Pipeline Details
  \begin{itemize}[leftmargin=2.5em, labelwidth=2.5em, labelsep=0em,
                  align=left, itemsep=1pt]
    \item[C.1] Visual Processing
    \item[C.2] Text Grounding
    \item[C.3] Question Generation
  \end{itemize}
  \item[\textbf{D.}] Additional Analysis of VLM Capabilities
  \begin{itemize}[leftmargin=2.5em, labelwidth=2.5em, labelsep=0em,
                  align=left, itemsep=1pt]
    \item[D.1] Human Expert Performance
    \item[D.2] Has GPT-5.4 Closed the Attribution Gap?
    \item[D.3] VLM Limitations Revealed by the Construction Framework
    \item[D.4] Interventional Validation of the Error Taxonomy
  \end{itemize}
  \item[\textbf{E.}] Additional Results
  \begin{itemize}[leftmargin=2.5em, labelwidth=2.5em, labelsep=0em,
                  align=left, itemsep=1pt]
    \item[E.1] Agreement Metric
    \item[E.2] Sample Balance
    \item[E.3] Agreement Across Judge Models
    \item[E.4] Ablation Replication Across Model Families
    \item[E.5] Rejected Question Analysis
    \item[E.6] VLM Error Case Studies
    \item[E.7] Rejected Question Examples
    \item[E.8] Accuracy Uncertainty Across Source Papers
    \item[E.9] Accuracy by Reasoning Type
  \end{itemize}
  \item[\textbf{F.}] Evidence-Grounded Attribution
  \item[\textbf{G.}] SMILES-First Pipeline
  \item[\textbf{H.}] Prompt Templates
  \item[\textbf{I.}] Human Evaluation Platform
  \item[\textbf{J.}] Benchmark Question Examples
  \item[\textbf{K.}] Design Considerations and Future Work
\end{itemize}

\vspace{1em}

\section{Extended Related Work}
\label{sec:related}

\noindent \textbf{Benchmarks for Chemical and Scientific Reasoning.}
A broad literature evaluates LLMs and VLMs on chemistry, but most benchmarks target prediction or retrieval rather than explanation of observed chemical effects. This distinction matters because a model can produce a correct product or pathway without identifying which structural feature actually drives the effect~\citep{chembench2024}.
Broad scientific benchmarks such as GPQA~\citep{rein2024gpqa}, SuperGPQA~\citep{du2025supergpqa}, and SciBench~\citep{wang2023scibench} evaluate advanced reasoning across disciplines, including chemistry, but they do not involve molecular diagrams or reaction schemes. Chemistry-focused benchmarks such as ChemLLMBench~\citep{guo2023ChemLLMBench}, ChemBench~\citep{chembench2024}, and ChemEval~\citep{huang2024chemeval} cover property prediction, structure elucidation, and name conversion~\citep{chemiq2024,moleculeqa2023,xie2025qcbench,guo2024molpuzzle}, while ChemCoTBench~\citep{chemcotbench2024}, oMeBench~\citep{omebench2024}, and SUPERChem~\citep{zhao2025superchem} focus on reaction outcome prediction and pathway generation. Literature-grounded chemistry QA benchmarks~\citep{moltextqa2024,chemrxivquest2024,scholarchemqa2024,chemlitqa2024} derive questions from papers, but they mainly evaluate factual retrieval. Across these benchmarks, the ability to attribute an observed chemical effect to the most responsible visible structural feature remains largely unmeasured. \ourbenchmark{} addresses this gap.

\noindent \textbf{Multimodal Chemistry Information Extraction and Literature-Based Benchmarks.}
A related line of work studies multimodal chemistry information extraction from scientific documents. These systems have progressed from text-based entity recognition~\citep{chemdataextractor2016} to multimodal pipelines~\citep{openchemie2024,mermaid2024,multimodalsearch2025} that recover molecular structures, reaction records, and associated text from papers. However, such pipelines typically retrieve all text linked to a molecule or reaction, without distinguishing explanatory evidence from procedures, observations, or characterization details. Existing quality control methods verify either the chemical validity of extracted structures~\citep{multimodalsearch2025} or the entity alignment between images and text~\citep{mermaid2024}, but they do not validate whether the retrieved text supports a structure-to-effect explanation. Literature-based chemistry benchmarks~\citep{moltextqa2024,chemrxivquest2024,scholarchemqa2024,chemlitqa2024} often rely on LLM-generated answers, which introduces an additional risk of hallucinated rationales~\citep{chemlitqa2024}. Agentic pipelines such as AutoConverter~\citep{zhang2025automated} and MicroVQA~\citep{burgess2025microvqa} convert human-written question--answer pairs into multiple-choice questions. Our framework instead validates explanatory evidence and derives each correct answer directly from the source paper.

\section{Representative Benchmark Example}
\label{sec:appendix_example}

Fig.~\ref{fig:appendix_sample} shows a complete \ourbenchmark{} question. Given a molecular structure image and an observed effect (high durability and high reaction quantum yield), the model must select the explanation that most directly accounts for the effect. Each of the four candidate explanations identifies a different structural feature visible in the depicted Re(I)--porphyrin complex and describes how it leads to the observed effect. The correct answer~(C) is grounded in the source paper evidence; the three distractors cite other features genuinely present in the structure---the bromide ligand (A), the carbonyl ligands (B), and the Zn(II)-metalated N4 core (D)---with plausible but incorrect mechanisms.

\section{Construction Pipeline Details}
\label{sec:impl_details}

This section provides the implementation details of the three construction stages described in Sections~\ref{sec:visual}--\ref{sec:synthesis}.

\subsection{Visual Processing}
\label{sec:impl_visual}

\noindent \textbf{Figure classification and segmentation.} A ResNet-50 classifier~\citep{he2016resnet} assigns each figure to one of four categories: single molecule, multiple molecules, reaction scheme, or non-chemistry. Non-chemistry figures, such as spectra, plots, and experimental images, are discarded. The remaining chemistry figures are then segmented into individual visual instances. Multi-molecule figures are processed with DECIMER Segmentation~\citep{rajan2021decimer}, whereas reaction schemes are decomposed into reactant, product, and condition components using RxnScribe~\citep{qian2023rxnscribe}. Single-molecule figures require no segmentation.

\noindent \textbf{Identifier extraction.} Chemistry papers typically annotate compounds with short identifiers that recur throughout the body text. MolCoref~\citep{openchemie2024} recovers them for only $\sim$27\% of molecule figures and $\sim$15\% of reaction schemes in our evaluation. Its extraction assumes regular grid-style layouts, whereas real paper figures often contain arrows, annotations, and mixed content, so many structures receive either no identifier or one belonging to a neighboring compound. We therefore extract the identifier directly from the figure with a VLM (GPT-4o-mini). The target structure is highlighted with a red bounding box in the original figure, and the VLM returns the corresponding identifier.

\noindent \textbf{Structure validation.} After segmentation and identifier extraction, the quality control agent verifies that each candidate region is a valid 2D chemical structure. This step removes 3D renderings, crystal packing diagrams, and schematic illustrations that may survive the upstream classifier.

\subsection{Text Grounding}
\label{sec:impl_text}

\noindent \textbf{Identifier-based retrieval.} For each structure, the text grounding agent forms a retrieval package consisting of the identifier, the structure image, the figure caption, and in-text figure references such as ``as shown in Fig.~3''. An LLM (GPT-4o-mini) then searches the full paper for sentences that discuss the target structure. Matching requires boundary-aware handling of short alphanumeric identifiers. For instance, identifier ``1'' should match ``compound 1'' but not the substring ``DFA1''.

\noindent \textbf{Context expansion.} Explanatory evidence in chemistry papers often spans adjacent sentences: one sentence states the observed effect, while a neighboring sentence provides the mechanism. For each retrieved sentence, the agent locates the containing paragraph and selects semantically related surrounding sentences with an LLM, forming an expanded evidence block. If any sentence in that block contains explanatory content, the entire block is retained, and partially explanatory sentences within it are kept as supporting context rather than as the basis for a question. Section~\ref{sec:ablation} quantifies the gain from this step.

\subsection{Question Generation}
\label{sec:impl_qgen}

\noindent \textbf{Explanatory decomposition.} Given a validated pair of structure image and explanatory text, an LLM (GPT-5.1) decomposes the explanatory text into three structured fields: (1) Effect: the observed effect, described without naming any structural feature (e.g., ``resists oxidative degradation''); (2) Feature: the structural feature identified in the source paper as the primary cause (e.g., ``bulky tert-butyl groups at ortho/para positions''); and (3) Mechanism: how the feature leads to the effect (e.g., ``steric shielding around phenolic OH slows oxidative attack''). The same call also returns three to five other structural features that are visible in the image but are not the cause given in the source text. If the explanatory text is too vague to yield all three fields, the candidate is discarded rather than completed by the model.

\noindent \textbf{Question and distractor generation.} The question text is generated from the Effect field only. The LLM does not receive the Feature or Mechanism fields during question generation, preventing the correct answer from leaking into the question text. The correct answer is then constructed from Feature + Mechanism. For the three distractors, the LLM draws on the other visible features identified during decomposition and pairs each with a chemically plausible mechanism for the same effect. To ensure faithfulness, the correct answer's mechanism must trace to the source paper, and the LLM is not permitted to introduce mechanisms beyond what the paper provides. Appendix~\ref{sec:prompts} gives the full prompt templates, including the complete set of hedging terms excluded under the equal-confidence constraint, such as ``indirect'' and ``secondary''.

\section{Additional Analysis of VLM Capabilities}
\label{sec:addition_VLM_cap}

\subsection{Human Expert Performance}
\label{sec:human_expert}

Eight chemistry domain experts answer the same 600-question stratified subset used for quality validation (Section~\ref{sec:quality_validation}). Tab.~\ref{tab:human_expert} reports their accuracy by reasoning category and entry type. These results are a reference for answerability and difficulty, not a direct comparison to the model results in Tab.~\ref{tab:main_results}, which cover the full benchmark.

\begin{table}[t]
\centering
\resizebox{\columnwidth}{!}{%
\begin{tabular}{@{}l|c|ccc|cc@{}}
\toprule
\textbf{Human Experts} & Avg. & Mech. & Design & Cond. & Mol. & Rxn. \\
\midrule
Average & 73.3 & 71.0 & 75.9 & 72.8 & 72.2 & 74.4 \\
Best & 80.0 & 77.4 & 82.8 & 80.4 & 70.0 & 90.0 \\
Worst & 68.3 & 67.7 & 69.0 & 68.6 & 66.7 & 70.0 \\
\bottomrule
\end{tabular}%
}
\caption{Human expert accuracy (\%) on the 600-question stratified subset, reported for eight chemistry domain experts. Mech.: Mechanistic; Cond.: Condition; Mol.: molecule; Rxn.: reaction.}
\label{tab:human_expert}
\end{table}

\noindent \textbf{Blind-phase agreement.}
Because experts answer each question before they see the source evidence (Section~\ref{sec:quality_validation}), their answers test whether the labeled answer can be identified from the question and the depicted structure alone. At least five of the eight experts select the labeled answer for 94.0\% of the 600 questions (Tab.~\ref{tab:blind_agreement}). If a distractor were an equally valid answer, experts who miss the labeled answer would converge on it. On the 414 questions that two or three experts answer incorrectly, however, these experts select different distractors in 321 cases (77.5\%). The labeled answer is therefore the consensus of chemists who have not seen the source paper, and individual errors reflect the limits of each expert's knowledge, as Expert~A's 94.1\% accuracy within his own domain illustrates (Appendix~\ref{sec:fine_grained}). 


\begin{table}[t]
\centering
\small
\begin{tabular}{@{}lc@{}}
\toprule
\textbf{Metric} & \textbf{Value} \\
\midrule
Mean expert accuracy & 73.3\% \\
Labeled answer chosen by $\geq$5 of 8 experts & 94.0\% \\
Labeled answer is the most-voted option & 99.7\% \\
Mean vote lead over the next option & 4.45 \\
Median vote lead over the next option & 4 \\
\bottomrule
\end{tabular}
\caption{Agreement of the eight experts' blind answers with the labeled answer on the 600-question subset. Tab.~\ref{tab:consensus} gives the full distribution of expert votes.}
\label{tab:blind_agreement}
\end{table}

GPT-5.4's attribution errors occur mainly on questions that most experts answer correctly without the source paper (Tab.~\ref{tab:consensus}). Of its 140 errors, 125 fall on questions that at least five experts answer correctly, and attribution accounts for 72.0\% of these errors, compared with 71.4\% overall. These errors therefore reflect a limitation of the model rather than uncertainty in the labeled answers.

\begin{table}[t]
\centering
\small
\begin{tabular}{@{}l|c|c|c@{}}
\toprule
\textbf{Votes} & \textbf{Questions} & \textbf{GPT-5.4 errors} & \textbf{Attr. errors} \\
\midrule
8/8 & 50 (8.3\%) & 5 & 4 \\
7/8 & 100 (16.7\%) & 15 & 12 \\
6/8 & 210 (35.0\%) & 45 & 36 \\
5/8 & 204 (34.0\%) & 60 & 38 \\
$\leq$4/8 & 36 (6.0\%) & 15 & 10 \\
\midrule
$\geq$5/8 & 564 (94.0\%) & 125 & 90 \\
Overall & 600 & 140 & 100 \\
\bottomrule
\end{tabular}
\caption{Distribution of blind expert votes for the labeled answer on the 600-question subset, with GPT-5.4 errors and the subset of these errors classified as attribution failures (Section~\ref{sec:error_analysis}). Votes: number of the eight experts selecting the labeled answer. Attr.: attribution.}
\label{tab:consensus}
\end{table}

\subsection{Has GPT-5.4 Closed the Attribution Gap?}
\label{sec:fine_grained}
 
On the full 1,180-question benchmark, GPT-5.4 reaches 78.6\% accuracy. On the 600-question stratified subset (Section~\ref{sec:quality_validation}), it reaches 76.7\%, exceeding the human expert average (73.3\%) and nearly matching the best individual expert (80.0\%). Does this mean current VLMs have reached human-level explanatory attribution? To investigate, we move beyond overall accuracy and compare the error profiles of GPT-5.4 and a human expert on this subset, categorizing each incorrect response into perception error, knowledge gap, and attribution error following the taxonomy in Section~\ref{sec:error_analysis}. We select Expert~A, a faculty member with 15+ years of experience specializing in organic and medicinal chemistry (71.7\% overall accuracy). During evaluation, Expert~A annotated each question with his confidence level and whether the topic fell outside his primary research area, allowing us to separate domain knowledge gaps from attribution errors.
 
\noindent \textbf{The expert's advantage: attribution depth.}
We examine how Expert~A performs on GPT-5.4's 140 error questions (Tab.~\ref{tab:expert_on_gpt_errors}). Expert~A answers 71.4\% correctly, resolving all knowledge gaps and perception errors, and 60\% of attribution errors. Of the remaining attribution errors, 75\% are on questions outside his expertise. Fig.~\ref{fig:case_human_wins} illustrates a representative case.
 
\begin{table}[h]
  \centering
  \small
  \renewcommand\arraystretch{1.0}
  \setlength{\tabcolsep}{15pt}
  \resizebox{\linewidth}{!}{%
  \begin{tabular}{l|c|c}
    \toprule
    \textbf{Error Type} & \textbf{Count} & \textbf{Expert~A Correct} \\
    \midrule
    Attribution Error & 100 & 60 \\
    Knowledge Gap     & 30  & 30 \\
    Perception Error  & 10  & 10 \\
    \midrule
    \textbf{Total}    & \textbf{140} & \textbf{100} \\
    \bottomrule
\end{tabular}
}
\caption{Expert~A's accuracy on the 140 questions GPT-5.4 answers incorrectly, grouped by error type (Section~\ref{sec:error_analysis}).}
\label{tab:expert_on_gpt_errors}
\end{table}

\noindent \textbf{The model's advantage: domain breadth.}
Conversely, of the 130 questions that GPT-5.4 answers correctly but Expert~A does not, 120 (92\%) fall outside Expert~A's domain, suggesting that these errors reflect unfamiliarity with specific chemistry subdomains rather than a failure in explanatory attribution. Fig.~\ref{fig:case_gpt_wins} shows a representative case.
 
\noindent \textbf{Decomposing overall accuracy.}
The findings above suggest that comparable overall accuracy masks different capability profiles. Tab.~\ref{tab:domain_split} quantifies this by splitting the 600 questions by whether Expert~A annotated the question as within or outside his expertise.
 
\begin{table}[t]
  \centering
  \small
  \renewcommand\arraystretch{1.0}
  \setlength{\tabcolsep}{11.5pt}
  \resizebox{\linewidth}{!}{%
  \begin{tabular}{l|cc|c}
    \toprule
    \textbf{Subset} & \textbf{Expert~A} & \textbf{GPT-5.4} & $\Delta$ \\
    \midrule
    Within-domain    & 94.1 & 70.6 & \textcolor{green!60!black}{$\uparrow$23.5} \\
    Out-of-domain    & 42.3 & 84.6 & \textcolor{red!70!black}{$\downarrow$42.3} \\
    \midrule
    Overall          & 71.7 & 76.7 & \textcolor{red!70!black}{$\downarrow$5.0} \\
    \bottomrule
\end{tabular}
}
\caption{Accuracy (\%) of Expert~A and GPT-5.4 on within-domain and out-of-domain subsets ($n$=340 and $n$=260 respectively), defined by Expert~A's self-annotations. $\Delta$: Expert~A accuracy minus GPT-5.4 accuracy.}
\label{tab:domain_split}
\end{table}
 
On questions within his expertise, Expert~A achieves 94.1\%, far exceeding GPT-5.4's 70.6\%. His overall accuracy (71.7\%) is reduced by out-of-domain questions (42.3\%), on which GPT-5.4 maintains 84.6\%. The answer to the opening question is clear: GPT-5.4 has not closed the attribution gap. Its competitive overall accuracy reflects broad domain coverage, not stronger explanatory attribution. The key direction for improving VLM performance on \ourbenchmark{} is not broader domain knowledge, but stronger explanatory attribution, the ability to determine which structural feature is most directly responsible for an observed effect.

\subsection{VLM Limitations Revealed by the Construction Framework}
\label{sec:pipeline_limitations}

The multi-agent framework relies on VLMs at multiple stages of the construction process, from aligning structures to text to evaluating explanatory quality. Analyzing where VLMs fail during construction reveals specific capability gaps of current models.

\noindent \textbf{Figure--text alignment.}
Chemistry papers often contain structural analogs that differ by a single substituent or functional group. Aligning each structure in a figure to its discussion in the text requires distinguishing these variants. Without identifier-based retrieval, the VLM achieves only 50\% entity match (Tab.~\ref{tab:ablation}): half of the retrieved text refers to a different compound rather than the target molecule or reaction. This indicates that current VLMs cannot reliably distinguish structural analogs from images and align them to the corresponding text in the paper.

\noindent \textbf{Distinguishing explanatory from descriptive text.}
In chemistry papers, most text that mentions a molecule or reaction is procedural or descriptive rather than explanatory (Section~\ref{sec:text}). Identifying explanatory sentences is the largest bottleneck in the construction framework: of 9,670 structures with retrieved text, only 1,264 (13.1\%) contain at least one explanatory sentence (Section~\ref{sec:construction}). Most discarded cases are partially explanatory: they identify a relevant structural feature and describe an observed effect, but do not provide the connecting mechanism, as in ``the bulky substituent improves selectivity'' without explaining how steric shielding leads to this outcome. Other discarded cases discuss a mechanism without linking it to any visible feature, as in PMC9719906 (chloroquine), where the retrieved text describes antiviral mechanisms only. This task is also difficult for current VLMs when explanatory content spans adjacent sentences, for example when the effect and the mechanism appear in different sentences of the same paragraph. On the 100-paper subset, 40\% of explanatory sentences are missed when sentences are evaluated in isolation rather than in context, which context expansion recovers (Tab.~\ref{tab:ablation}). This limitation is consistent with the evaluation finding in Section~\ref{sec:error_analysis}: on the benchmark, VLMs struggle most with explanatory attribution, determining which structural feature is most directly responsible for an observed effect. In the construction framework, VLMs similarly struggle with identifying explanatory evidence. Improving VLM capabilities in explanatory reasoning would therefore benefit not only benchmark performance but also the ability to extract explanatory evidence from scientific literature.

\subsection{Interventional Validation of the Error Taxonomy}
\label{sec:intervention}

Section~\ref{sec:error_analysis} assigns error types from reasoning traces. To test whether these types reflect where models fail, we add expert-verified oracle information to the prompt. If the error types reflect where models fail, structural descriptions should correct perception errors and chemical knowledge should correct knowledge gaps, whereas attribution errors, whose traces already show correct perception and knowledge, should persist when both are supplied. We evaluate GPT-5.4 and Qwen3-VL-8B on the questions they answer incorrectly in the 600-question subset (140 and 300, of which 100 are answered incorrectly by both), together with 100 control questions answered correctly by both models, for 440 distinct questions. Each question is evaluated three times under four conditions and scored by majority vote: a rerun of the original prompt (C0), descriptions of the structural features referenced by the options (C1), chemical knowledge linking each option's feature to the effect (C2), and both (C3). Two-sided exact McNemar tests compare each condition with C0.

\noindent \textbf{Oracle construction and verification.}
Claude Sonnet 4.6 drafts one C1 structural description for each of the 440 questions from the molecular or reaction image and the four options, without access to the labeled answer. A fixed template requires one description of each structural feature referenced by an option and prohibits comparative terms such as ``stronger,'' ``dominant,'' or ``more important.'' A chemistry expert blind to the labeled answer reviews all 440 descriptions for structural correctness, localization, completeness, and option neutrality, accepting 421 (95.7\%) as written and revising 19 (4.3\%). A second expert independently reviews all 440 descriptions and revises three more. Both experts judge every description to be neutral across options. For C2, each option is paired with a short, expert-verified statement of the chemical knowledge linking its feature to the effect, including the conditions under which the underlying principle applies. These statements are matched in level of detail across options and do not rank the options, leaving the judgment of which feature is primarily responsible to the model.

\noindent \textbf{Results.}
The oracle information corrects perception errors and knowledge gaps but leaves attribution errors largely uncorrected (Tab.~\ref{tab:intervention}). Structural descriptions raise the correction rate of perception errors, the percentage answered correctly after the intervention, from 20.0\% under C0 to 80.0\% for GPT-5.4 and from 15.0\% to 70.0\% for Qwen3-VL-8B. Chemical knowledge raises that of knowledge gaps from 16.7\% to 70.0\% and from 13.3\% to 65.6\%. Both effects are significant, and neither type of information significantly affects the other error types (Tab.~\ref{tab:intervention_mcnemar}). For attribution errors, supplying both raises the correction rate from 14.0\% to 19.0\% and from 12.0\% to 16.0\%, and no condition has a significant effect. Accuracy on the control questions stays within one point of C0 under every condition (Tab.~\ref{tab:intervention_control}), so the oracle information does not alter answers that are already correct. The error types assigned from reasoning traces therefore reflect where the models fail, and attribution errors arise from misjudging which feature is primarily responsible rather than from undetected failures of perception or knowledge.

\begin{table*}[t]
\centering
\small
\begin{tabular}{llccccc}
\toprule
\textbf{Model} & \textbf{Error type} & \textbf{N} & \textbf{C0} & \textbf{C1: Structure} & \textbf{C2: Knowledge} & \textbf{C3: Both} \\
\midrule
\multirow{3}{*}{GPT-5.4}
& Perception  & 10  & 20.0 (2/10)   & \textbf{80.0 (8/10)}  & 20.0 (2/10)           & \textbf{90.0 (9/10)} \\
& Knowledge   & 30  & 16.7 (5/30)   & 23.3 (7/30)           & \textbf{70.0 (21/30)} & \textbf{76.7 (23/30)} \\
& Attribution & 100 & 14.0 (14/100) & 17.0 (17/100)         & 16.0 (16/100)         & 19.0 (19/100) \\
\midrule
\multirow{3}{*}{Qwen3-VL-8B}
& Perception  & 60  & 15.0 (9/60)   & \textbf{70.0 (42/60)} & 16.7 (10/60)          & \textbf{76.7 (46/60)} \\
& Knowledge   & 90  & 13.3 (12/90)  & 17.8 (16/90)          & \textbf{65.6 (59/90)} & \textbf{71.1 (64/90)} \\
& Attribution & 150 & 12.0 (18/150) & 14.7 (22/150)         & 14.0 (21/150)         & 16.0 (24/150) \\
\bottomrule
\end{tabular}
\caption{Correction rates (\%) of originally incorrect questions under each condition, with counts in parentheses. C0 reruns the original prompt, C1 adds structural descriptions, C2 adds chemical knowledge, and C3 adds both. Bold marks a significant difference from C0 (two-sided exact McNemar test, $p<0.05$).}
\label{tab:intervention}
\end{table*}

\begin{table*}[t]
\centering
\small
\begin{tabular}{llccc}
\toprule
\textbf{Model} & \textbf{Error type} & \textbf{C1 vs.\ C0} & \textbf{C2 vs.\ C0} & \textbf{C3 vs.\ C0} \\
\midrule
\multirow{3}{*}{GPT-5.4}
& Perception  & $b=6,c=0,\;p=.031$  & $b=1,c=1,\;p=1.000$  & $b=7,c=0,\;p=.016$ \\
& Knowledge   & $b=4,c=2,\;p=.688$  & $b=18,c=2,\;p<.001$  & $b=20,c=2,\;p<.001$ \\
& Attribution & $b=8,c=5,\;p=.581$  & $b=7,c=5,\;p=.774$   & $b=10,c=5,\;p=.302$ \\
\midrule
\multirow{3}{*}{Qwen3-VL-8B}
& Perception  & $b=35,c=2,\;p<.001$ & $b=5,c=4,\;p=1.000$  & $b=39,c=2,\;p<.001$ \\
& Knowledge   & $b=8,c=4,\;p=.388$  & $b=49,c=2,\;p<.001$  & $b=54,c=2,\;p<.001$ \\
& Attribution & $b=9,c=5,\;p=.424$  & $b=8,c=5,\;p=.581$   & $b=11,c=5,\;p=.210$ \\
\bottomrule
\end{tabular}
\caption{Two-sided exact McNemar tests comparing each condition with C0. $b$: questions correct under the condition but incorrect under C0. $c$: questions correct under C0 but incorrect under the condition.}
\label{tab:intervention_mcnemar}
\end{table*}

\begin{table}[t]
\centering
\small
\begin{tabular}{lcccc}
\toprule
\textbf{Model} & \textbf{C0} & \textbf{C1} & \textbf{C2} & \textbf{C3} \\
\midrule
GPT-5.4     & 96.0 & 97.0 & 96.0 & 97.0 \\
Qwen3-VL-8B & 92.0 & 93.0 & 92.0 & 93.0 \\
\bottomrule
\end{tabular}
\caption{Accuracy (\%) on 100 control questions that both models originally answer correctly. No oracle condition differs significantly from C0 (two-sided exact McNemar tests, all $p>0.05$).}
\label{tab:intervention_control}
\end{table}

\section{Additional Results}
\label{sec:additional_results}

\subsection{Agreement Metric}
\label{sec:agreement_metric}
We report Gwet's AC$_2$ with quadratic weights over the full category set $[1,2,3,4,5]$~\citep{gwet2014handbook}. Quality ratings in Tab.~\ref{tab:agreement} cluster in the upper part of the Likert scale, which triggers the prevalence paradox~\citep{wongpakaran2013comparison,feinstein1990high}: when raters agree on most items and those items concentrate in one category, chance-corrected coefficients such as weighted $\kappa$ and Krippendorff's $\alpha$ subtract an inflated chance term and can report low agreement despite near-identical ratings. AC$_2$ estimates the chance term from the observed rating distribution and remains stable in this regime. Recent NLP evaluation studies likewise find AC$_2$ stable under skewed rating distributions~\citep{pradhan2025llmjudge,james2026counting,liu2026retell}. Confidence intervals follow from the analytic standard errors in Tab.~\ref{tab:agreement}; the interval for the average uses the mean of the four per-dimension standard errors. The largest standard error for human-LLM AC$_2$ is 0.022, corresponding to an approximate 95\% confidence-interval half-width of 0.043. The judge's scores agree with blind expert scores on rejected questions as well as accepted ones (Appendix~\ref{sec:failure_table}).

\subsection{Sample Balance}
\label{sec:sample_balance}
Tab.~\ref{tab:sample_balance} compares the judge's scores for the 600-question subset against the full benchmark. Holm--Bonferroni correction over the four comparisons compares the smallest $p$-value against $0.05/4 = 0.0125$ and stops if it does not pass. The smallest value is 0.021, so none of the four comparisons is statistically significant at the 5\% level.

\begin{table}[t]
\centering
\resizebox{0.92\columnwidth}{!}{%
\begin{tabular}{@{}l|c|c|c|c@{}}
\toprule
\textbf{Dimension} & Benchmark & Subset & $\Delta$ & $p$ \\
\midrule
Distractor Quality & 4.447 & 4.450 & $+$0.003 & 0.720 \\
Meaningfulness     & 4.447 & 4.470 & $+$0.023 & 0.400 \\
Answer Unambiguity & 4.480 & 4.460 & $-$0.020 & 0.440 \\
Struct.\ Necessity & 4.505 & 4.630 & $+$0.125 & 0.021 \\
\bottomrule
\end{tabular}%
}
\caption{Judge scores on the 1--5 Likert scale for the full benchmark and the 600-question expert subset. $\Delta$: Subset minus Benchmark. $p$: two-sided Mann--Whitney U test.}
\label{tab:sample_balance}
\end{table}

\subsection{Agreement Across Judge Models}
\label{sec:judge_robustness}

To assess whether the screening depends on the choice of judge, the 600-question expert-rated subset is scored by three judges from different model families, GPT-5.2, Claude-Opus-4-7, and Gemini-3.1-Pro, each following the protocol of Section~\ref{sec:quality_validation}. Tab.~\ref{tab:judge_robustness} reports human--LLM agreement (Gwet's AC$_2$) for each judge. All three reach comparably high agreement with the human experts (average AC$_2$ of 0.919, 0.938, and 0.949, respectively), indicating that the agreement reflects the quality criteria rather than the choice of judge.

\begin{table}[t]
\centering
\resizebox{\columnwidth}{!}{%
\begin{tabular}{@{}l|cccc|c@{}}
\toprule
\textbf{Judge (H--L AC$_2$)} & Distr. & Mean. & Unamb. & Struct. & \textbf{Avg.} \\
\midrule
GPT-5.2 (orig.) & 0.948 & 0.936 & 0.929 & 0.861 & 0.919 \\
Claude-Opus-4-7 & 0.933 & 0.939 & 0.926 & 0.952 & 0.938 \\
Gemini-3.1-Pro & 0.952 & 0.940 & 0.932 & 0.973 & 0.949 \\
\bottomrule
\end{tabular}%
}
\caption{Human--LLM agreement (Gwet's AC$_2$) on the 600-question expert subset for three judges from different model families. Distr.: Distractor Quality; Mean.: Meaningfulness; Unamb.: Answer Unambiguity; Struct.: Structure Necessity.}
\label{tab:judge_robustness}
\end{table}

\subsection{Ablation Replication Across Model Families}
\label{sec:ablation_claude}

The ablation is repeated on the same 100-paper subset with Claude-Sonnet-4-6 at every stage, under the metrics and configurations of Section~\ref{sec:ablation}. Tab.~\ref{tab:ablation_claude} reports the result. Absolute counts shift with the model, but removing image-first construction, replacing VLM-based identifier extraction, and removing context expansion each degrade the corresponding metric, reproducing the pattern in Tab.~\ref{tab:ablation}.

\begin{table}[t]
\centering
\resizebox{\columnwidth}{!}{%
\begin{tabular}{@{}l|c|c|c@{}}
\toprule
\textbf{Config (Claude-Sonnet-4-6)} & Struct. & Ent. Match & Expl. \\
\midrule
Full pipeline & 489 & 78\% & 41 \\
\quad w/o image-first (SMILES) & 154 & --- & 3 \\
\quad w/o VLM identifier & 489 & 45\% & 31 \\
\quad w/ MolCoref identifier & 489 & 36\% & 27 \\
\quad w/o context expansion & 489 & 79\% & 24 \\
\bottomrule
\end{tabular}%
}
\caption{Ablation on the same 100-paper subset as Tab.~\ref{tab:ablation}, repeated with Claude-Sonnet-4-6. Struct.: individual molecular or reaction images retained after classification and segmentation. Ent.\ Match: fraction of retrieved text correctly referring to the target molecule or reaction. Expl.: explanatory sentences that qualify for question generation.}
\label{tab:ablation_claude}
\end{table}

\subsection{Rejected Question Analysis}
\label{sec:failure_table}

The LLM judge rejects 84 candidate questions during post-generation screening (Section~\ref{sec:quality_validation}), 83 with average scores between 3.0 and 4.0 and one below 3.0. The eight domain experts evaluate 60 of them on the four quality dimensions, blind to the judge's decisions. Tab.~\ref{tab:failure} reports the scores. The mean expert score decreases from 4.53 on accepted questions to 3.68 on rejected questions, and the mean judge score decreases correspondingly from 4.50 to 3.68. The judge therefore distinguishes accepted from rejected candidates consistently with blind expert assessment.

\noindent \textbf{Rejection causes.}
We attribute each rejection to the dimension with the lowest judge score, and Tab.~\ref{tab:failure} reports the share of rejections attributed to each dimension. Structure Necessity accounts for 76.7\% of rejections: these questions receive a mean expert score of 2.70 on this dimension, compared with 4.75 for accepted questions, because they can be answered from the option text or general chemistry without the depicted structure (Fig.~\ref{fig:failure_necessity}). Correct Answer Unambiguity accounts for another 20.0\%, typically because another option offers an equally defensible explanation (Fig.~\ref{fig:failure_ambiguity}). Distractor Quality and Question Meaningfulness are rarely the primary cause (3.3\% and 0.0\%), indicating that the generation constraints already address them.

\begin{table}[t]
  \centering
  \small
  \setlength{\tabcolsep}{13pt}
  \resizebox{\linewidth}{!}{%
  \begin{tabular}{l|cc|c}
    \toprule
    \textbf{Dimension} & $\bar{H}$ & $\bar{L}$ & \textbf{Cause (\%)} \\
    \midrule
    Structure Necessity & 2.70 & 2.69 & 76.7 \\
    Answer Unambiguity & 3.89 & 3.93 & 20.0 \\
    Distractor Quality & 3.90 & 3.85 & 3.3 \\
    Meaningfulness & 4.23 & 4.23 & 0.0 \\
    \midrule
    Overall & 3.68 & 3.68 & --- \\
    \bottomrule
  \end{tabular}
}
  \caption{Quality scores for 60 rejected questions ($n{=}60$ from 84 rejected candidates, 8 human experts). $\bar{H}$: mean human expert score (1--5); $\bar{L}$: mean LLM judge score. Cause~(\%): dimension with the lowest LLM score, indicating the primary reason for rejection.}
  \label{tab:failure}
\end{table}

\subsection{VLM Error Case Studies}
\label{sec:error_cases}

We present one representative example of each error type from the manual analysis in Section~\ref{sec:error_analysis}. Fig.~\ref{fig:attribution_example} illustrates an attribution error, Fig.~\ref{fig:case_perception} a perception error, and Fig.~\ref{fig:case_knowledge} a knowledge gap.

\subsection{Rejected Question Examples}
\label{sec:construction_failures}

We present representative examples of questions rejected during post-generation screening (Section~\ref{sec:quality_validation}), illustrating the most common rejection causes identified in Appendix~\ref{sec:failure_table}. Fig.~\ref{fig:failure_necessity} shows a structure necessity rejection and Fig.~\ref{fig:failure_ambiguity} shows an answer unambiguity rejection.

\subsection{Accuracy Uncertainty Across Source Papers}
\label{sec:paper_level}

\begin{table*}[t]
\centering
\small
\begin{tabular}{@{}l|c|c|c@{}}
\toprule
\textbf{Model} & \textbf{Accuracy} & \textbf{Paper-level 95\% CI} & \textbf{One question per paper} \\
\midrule
GPT-5.4 & 78.6 & [75.7, 81.4] & 78.4 [76.0, 80.9] \\
GPT-5 & 77.4 & [74.4, 80.2] & 77.2 [74.9, 79.6] \\
GPT-4o & 68.1 & [65.0, 71.2] & 67.8 [65.3, 70.2] \\
Claude Sonnet 4.6 & 78.2 & [75.4, 80.9] & 78.1 [75.7, 80.7] \\
Gemini-3.1-Pro & 78.0 & [75.1, 80.7] & 77.9 [75.5, 80.4] \\
\midrule
Qwen2.5-VL 7B & 42.6 & [39.0, 46.2] & 42.5 [39.9, 45.2] \\
Qwen3-VL 8B & 52.5 & [48.9, 56.0] & 52.3 [49.6, 55.1] \\
Qwen3-VL 30B & 58.8 & [55.3, 62.2] & 58.7 [55.9, 61.6] \\
Phi-4-reasoning-vision 15B & 51.3 & [47.8, 54.8] & 51.0 [48.3, 53.8] \\
InternVL3.5 8B & 48.0 & [44.5, 51.4] & 47.8 [45.2, 50.4] \\
Janus-Pro 7B & 23.2 & [20.3, 26.3] & 23.3 [20.9, 25.6] \\
\midrule
ChemVLM 8B & 34.0 & [30.7, 37.3] & 33.8 [31.3, 36.3] \\
\bottomrule
\end{tabular}
\caption{Accuracy (\%) on the full benchmark with paper-level uncertainty. Paper-level 95\% CI: percentile interval from 10{,}000 bootstrap resamples of source papers. One question per paper: mean accuracy and 95\% range over 1{,}000 resamples that keep one randomly chosen question from each paper.}
\label{tab:paper_level}
\end{table*}

We examine two properties of the accuracy estimates: whether differences between models exceed sampling variation, and whether the estimates depend on papers that contribute more questions. Questions from the same source paper tend to share a chemical domain, drawing style, and terminology and are therefore not independent, so we resample papers rather than individual questions. Tab.~\ref{tab:paper_level} reports accuracy with 95\% confidence intervals from 10{,}000 paper-level bootstrap resamples, together with mean accuracy over 1{,}000 resamples that keep one randomly chosen question per paper.

Under paper-level resampling, GPT-5.4, Claude Sonnet 4.6, and Gemini-3.1-Pro are statistically indistinguishable: the 95\% confidence intervals of all their paired differences include zero (GPT-5.4 vs.\ Claude Sonnet 4.6: $+$0.4 [$-$2.5, 3.3]; GPT-5.4 vs.\ Gemini-3.1-Pro: $+$0.6 [$-$2.3, 3.6]; Claude Sonnet 4.6 vs.\ Gemini-3.1-Pro: $+$0.2 [$-$2.4, 2.9]). Consistent with this, each of the four leading proprietary models ranks first in at least 13\% of the one-question-per-paper resamples (GPT-5.4: 39.6\%, Claude Sonnet 4.6: 25.8\%, Gemini-3.1-Pro: 21.5\%, GPT-5: 13.1\%).

The estimates also do not depend on how many questions each paper contributes. Each of the 383 papers contributes 3.08 questions on average (median 3, maximum 8), and keeping one question per paper changes accuracy by at most 0.3 points for every model.

\subsection{Accuracy by Reasoning Type}
\label{sec:reasoning_type}

Design questions are the easiest category for all evaluated models and for the human experts. Mechanistic and Condition questions are more challenging, for example for Qwen3-VL-30B (54.7\% and 54.9\% vs.\ 65.0\% on Design; Tab.~\ref{tab:main_results}) and for the human experts (71.0\% on Mechanistic vs.\ 75.9\% on Design; Tab.~\ref{tab:human_expert}).

\section{Evidence-Grounded Attribution}
\label{sec:ega}

The error patterns in Section~\ref{sec:error_analysis} suggest two directions for model development. Some perception and knowledge errors could be reduced through tool-augmented inference, for example by integrating structure recognition models or chemical knowledge bases. Attribution errors cannot be resolved by structure recognition or knowledge retrieval alone, because the difficulty is not access to the structure or the chemistry but ranking competing explanations within a specific structural context. This makes attribution a suitable target for structured supervision: the decomposition used in \ourbenchmark{} (Effect, Feature, and Mechanism) exposes intermediate reasoning stages that could support hierarchical training or reward design. Such supervision would encourage models to first identify relevant structural evidence, then establish the mechanism, and only then decide which feature is primary. This section tests whether the same ordering can improve inference through prompting alone, without fine-tuning or external tools.

\noindent \textbf{Motivation and error pattern.}
The strongest open-source model trails the leading proprietary models by nearly 20 percentage points on the full benchmark (Tab.~\ref{tab:main_results}), leaving substantial room for improvement. The error analysis further classifies most of GPT-5.4's errors as attribution failures, in which the reasoning trace identifies the relevant structural features and applies the necessary chemical knowledge but selects the wrong feature as the primary explanation. To better understand the pattern, a chemistry expert examined the reasoning traces for the 100 attribution errors made by GPT-5.4 on the 600-question subset. In many cases, the model favored an explanation that is broadly common in chemistry, such as electron withdrawal by a carbonyl, over a less typical feature supported by the depicted structure, such as a nonclassical C--H$\cdots$O contact. This pattern suggests that generic chemical priors can override evidence specific to the structure under examination. We therefore use open-source models as a focused and reproducible test bed for evaluating whether this benchmark-derived insight can improve performance where the gap is largest.

\noindent \textbf{Method.}
Evidence-Grounded Attribution (EGA) requires the model to ground every candidate explanation in the depicted structure before selecting an answer. For each option, the model first identifies the visual evidence that would support the named feature and then explains how that feature could produce the observed effect. It is instructed to discount options that are chemically plausible in general but are not well supported by the specific structure. After comparing the four grounded explanations, the model selects the feature-mechanism relation that best accounts for the effect. EGA uses a single model call and requires no external tools, SMILES conversion, or fine-tuning.

\noindent \textbf{Results.}
We apply EGA to Qwen2.5-VL, Qwen3-VL-8B, and ChemVLM, covering two generations of a widely used open-source model family and one chemistry-specific model. All models are evaluated on the full benchmark using the same zero-shot, three-trial majority-vote protocol as the main experiments. EGA improves all three models by 3.1--5.3 percentage points (Tab.~\ref{tab:ega}), showing that the benchmark analysis yields an actionable inference-time strategy rather than only a descriptive error taxonomy. The gains remain moderate because EGA targets attribution, while the error analysis also identifies substantial perception and knowledge failures in these models. The strongest EGA result reaches 55.6\%, leaving considerable room for further improvement.

\begin{table}[t]
\centering
\setlength{\tabcolsep}{13pt}
\resizebox{0.8\linewidth}{!}{%
\begin{tabular}{@{}l|c|c|c@{}}
\toprule
\textbf{Model} & Baseline & EGA & $\Delta$ \\
\midrule
Qwen2.5-VL (7B) & 42.6 & 46.8 & $+$4.2 \\
Qwen3-VL (8B)   & 52.5 & 55.6 & $+$3.1 \\
ChemVLM (8B)    & 34.0 & 39.3 & $+$5.3 \\
\bottomrule
\end{tabular}%
}
\caption{Accuracy (\%) with Evidence-Grounded Attribution on the full benchmark under the protocol of Tab.~\ref{tab:main_results}. $\Delta$ denotes the absolute
percentage-point improvement.}
\label{tab:ega}
\end{table}

\section{SMILES-First Pipeline}
\label{sec:smiles_pipeline}

As described in Section~\ref{sec:visual}, the image-first pipeline retains all structures as images rather than converting them to SMILES strings. This section describes the SMILES-first alternative that was evaluated and abandoned due to the systematic loss of complex structures reported in Section~\ref{sec:ablation}.

In this alternative, segmented structures are converted to SMILES strings before text grounding. For single molecules, MolScribe~\citep{qian2023molscribe} converts the segmented image to a SMILES string. For multi-molecule figures, DECIMER Segmentation~\citep{rajan2021decimer} first segments the figure into individual molecules, each of which is converted to a SMILES string by MolScribe. For reaction schemes, RxnScribe~\citep{qian2023rxnscribe} segments the reaction into individual components and extracts SMILES for each molecular component. If SMILES extraction fails for any component, the entire reaction is discarded.

We apply round-trip quality control to verify extracted SMILES strings.The extracted SMILES is redrawn into a standardized structure image using RDKit. DECIMER then processes this redrawn image to produce a second SMILES string. Both the original and regenerated SMILES are converted to molecular graphs using NetworkX and compared via graph isomorphism. If the similarity is at least 0.85, the structure passes validation; otherwise it is discarded. For reactions, all molecular components must individually pass this validation. Section~\ref{sec:ablation} reports the quantitative comparison with the image-first pipeline. Fig.~\ref{fig:smiles_failures} illustrates the SMILES extraction and validation procedure together with two representative failure modes.

\section{Prompt Templates}
\label{sec:prompts}

This section provides the key prompt templates used in question generation (Section~\ref{sec:synthesis}) and LLM-as-a-Judge screening (Section~\ref{sec:quality_validation}). Fig.~\ref{fig:prompt_decomposition} shows the Stage 1 explanatory decomposition prompt, Fig.~\ref{fig:prompt_generation} shows the Stage 2 question and distractor generation prompt, and Fig.~\ref{fig:prompt_judge} shows the LLM-as-a-Judge screening prompt.

\section{Human Evaluation Platform}
\label{sec:eval_platform}

Fig.~\ref{fig:eval_platform_1} shows the blind answer phase, in which evaluators answer the question without access to the source paper text. Fig.~\ref{fig:eval_platform_2} shows the evidence review and scoring phase, in which evaluators review the paper evidence and score the four quality dimensions on a 1--5 Likert scale.


\begin{figure}[t]
  \centering
  \includegraphics[width=0.8\linewidth]{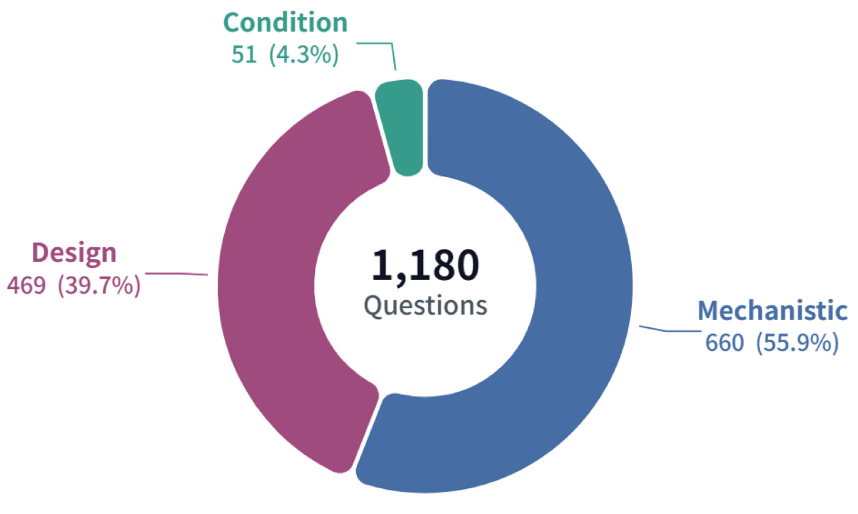}
  \caption{Composition of the 1,180-question benchmark sourced from 383 papers, distributed across three reasoning categories: Mechanistic (55.9\%), Design (39.7\%), and Condition (4.3\%).}
  \label{fig:distribution}
\end{figure}

\section{Benchmark Question Examples}
\label{sec:question_examples}

Fig.~\ref{fig:distribution} shows how the 1,180 questions are distributed across the three reasoning categories. We present representative \ourbenchmark{} questions across reasoning categories (Mechanistic, Design, Condition) and entry types (molecule, reaction). Each example shows the molecular or reaction image, the question, four candidate explanations, and the correct answer. Figs.~\ref{fig:example_mech_mol}--\ref{fig:example_cond_rxn} show molecule and reaction examples for each reasoning category.

\section{Design Considerations and Future Work}
\label{sec:limitations}

\noindent \textbf{Robustness to pretraining exposure.}
Because the source articles are published papers, some may appear in the pretraining data of the evaluated models. Several design choices prevent a model from exploiting memorized text: the questions and distractors are newly written rather than copied from the papers, and answer positions are randomized (Section~\ref{sec:synthesis}). The text-only control in Section~\ref{sec:visual_necessity} further shows that source-text or option recall alone cannot explain the strong image-based performance. To test whether recognizing a source paper or figure raises accuracy, we compare GPT-4o on two category-matched sets of 200 questions, drawn from papers published before and after its documented knowledge cutoff (October 1, 2023). Its accuracy is 68.5\% and 68.0\%, respectively, a difference of one question (95\% CI $[-8.6, 9.6]$ points), so prior exposure to the source papers does not measurably raise its accuracy. The automated pipeline can also generate future benchmark versions from papers published after a given model's training cutoff.

\noindent \textbf{Safeguards in LLM-based construction.}
The framework uses LLMs and VLMs for identifier extraction, text grounding, explanatory-value assessment, and question generation, and each of these steps is constrained by source evidence or checked by quality control. The correct feature-mechanism relation must be supported by evidence from the source paper, and the question-generation model cannot introduce a mechanism absent from that evidence (Section~\ref{sec:synthesis}). The anti-shortcut constraints control stylistic artifacts in the answer options, and quality control filters outputs before they advance to the next stage. Expert evaluation supports this design: accepted questions receive an average score of 4.53 out of 5 (Tab.~\ref{tab:agreement}), and every question in the expert subset passes the 4.0 acceptance threshold under expert scores. Future versions could further automate the explanatory-value assessment with a classifier trained on the accumulated labels.

\noindent \textbf{Model-agnostic construction.}
The construction is designed to keep the benchmark independent of the model used for question generation. Each correct feature-mechanism relation is taken from the source paper, and the generator (GPT-5.1) only rewrites this evidence into a question and options (Section~\ref{sec:synthesis}). The evaluation results confirm this property across model families. GPT-5.4, Claude Sonnet 4.6, and Gemini-3.1-Pro lie within 0.6 points of one another, Claude Sonnet 4.6 and Gemini-3.1-Pro both outperform GPT-5, and Claude Sonnet 4.6 achieves the highest Mechanistic accuracy (Tab.~\ref{tab:main_results}). Without the image, the three frontier models reach similar accuracies of 20.5--24.5\% (Tab.~\ref{tab:text_only}), so the option wording is equally informative across model families. Judges from three model families also reach comparable agreement with human experts (Appendix~\ref{sec:judge_robustness}), and the construction ablations yield the same ordering of design choices with Claude Sonnet 4.6 (Appendix~\ref{sec:ablation_claude}).

\noindent \textbf{Determining the most direct explanation.}
An observed chemical effect may involve several structural features, so each question asks which feature is primarily responsible. The correct feature-mechanism relation is grounded in the source paper, and distractors are drawn from other features visible in the same structure that the paper does not identify as primary (Section~\ref{sec:synthesis}). Experts who answer without the source paper select the labeled answer by majority on 94.0\% of the expert subset (Appendix~\ref{sec:human_expert}), so in most questions the explanation that the source paper identifies as primary is also the one that independent chemists identify. The Correct Answer Unambiguity check removes questions for which another option is equally defensible, accounting for 20.0\% of rejected questions (Tab.~\ref{tab:failure}). Accepted questions receive an average expert score of 4.43 out of 5 on this dimension (Tab.~\ref{tab:agreement}), and none of the 4,800 expert ratings falls at or below 3, the rubric level for genuine scientific debate about the primary cause (Fig.~\ref{fig:prompt_judge}). The LLM judge applies the same check to every candidate question and agrees closely with the experts on this dimension (AC$_2$=0.929), so the expert results extend to the screening of the full benchmark. Future versions could also retain the removed questions through graded or multi-answer evaluation, extending coverage to effects that several features explain jointly.

\noindent \textbf{Category composition.}
The category proportions follow the source literature. Condition questions account for 4.3\% of the benchmark (51 questions) because source papers rarely state explicitly why a structural feature requires particular reaction conditions. The construction framework can increase this share by targeting papers that discuss stoichiometry, selectivity, catalyst choice, and other condition-dependent effects.

\noindent \textbf{Source corpus.}
The source corpus is drawn from the PubMed Central Open Access Subset and restricted to journals with SJR $\geq 2.0$ (Section~\ref{sec:construction}). The Open Access Subset provides articles under licenses that permit text mining and derivative use, and the SJR threshold favors papers that discuss structure-to-effect relationships explicitly. The resulting corpus spans organic synthesis, materials chemistry, catalysis, and medicinal chemistry. The construction pipeline can be applied to additional repositories and chemistry subdomains while preserving the license of each source paper, evidence grounding, and quality control.


\onecolumn

\begin{figure}[t]
  \centering
  \includegraphics[width=0.9\textwidth]{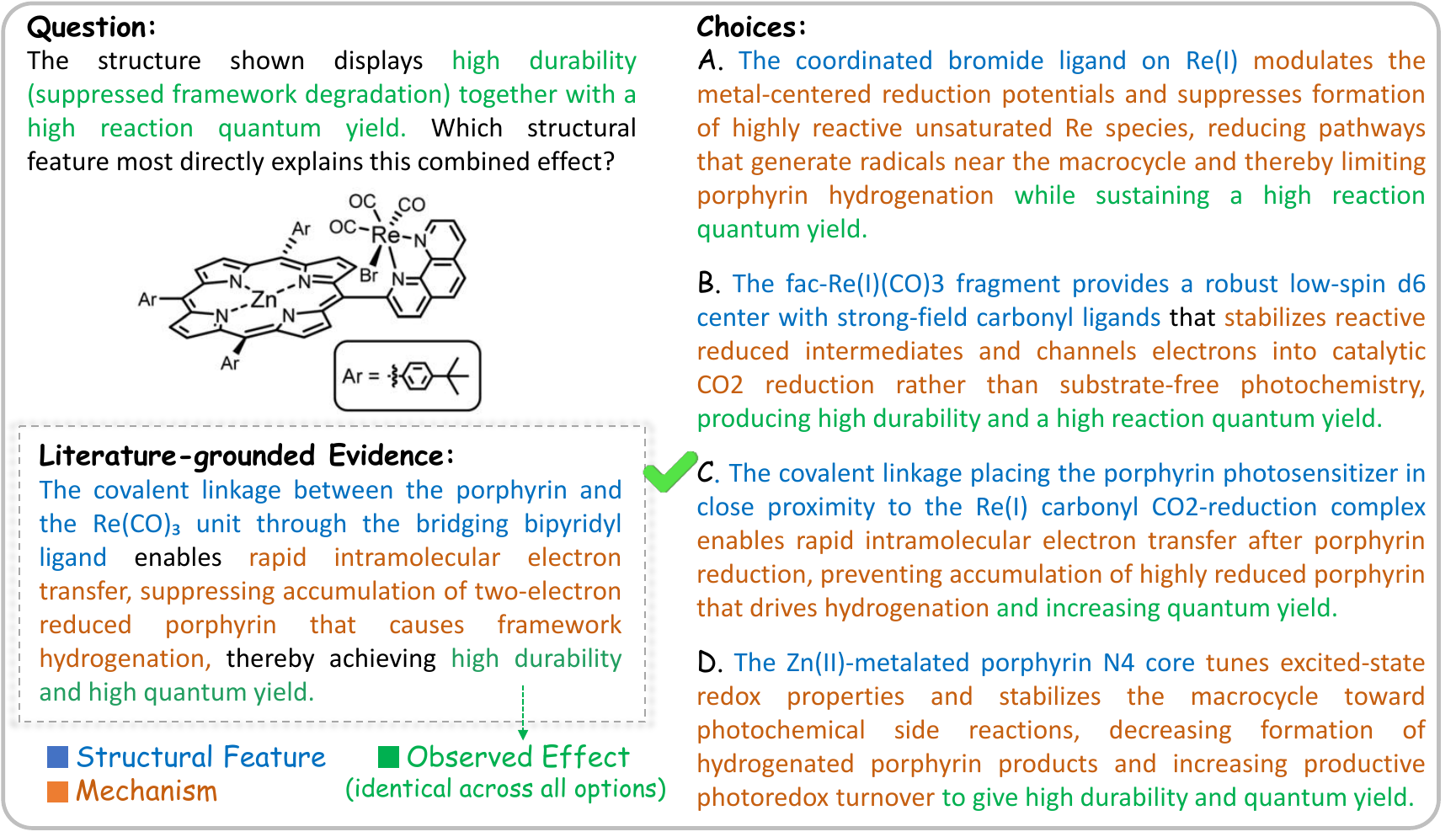}
  \caption{A complete annotated \ourbenchmark{} question: a molecular structure image, an observed effect (green, identical across all options), and four candidate explanations each identifying a different structural feature (blue) and mechanism (orange). The correct answer (C, checkmark) is grounded in the source paper evidence (dashed box). Distractors reference other structural features genuinely present in the molecule.}
  \label{fig:appendix_sample}
\end{figure}

\begin{figure}[h]
  \centering
  \includegraphics[width=\textwidth]{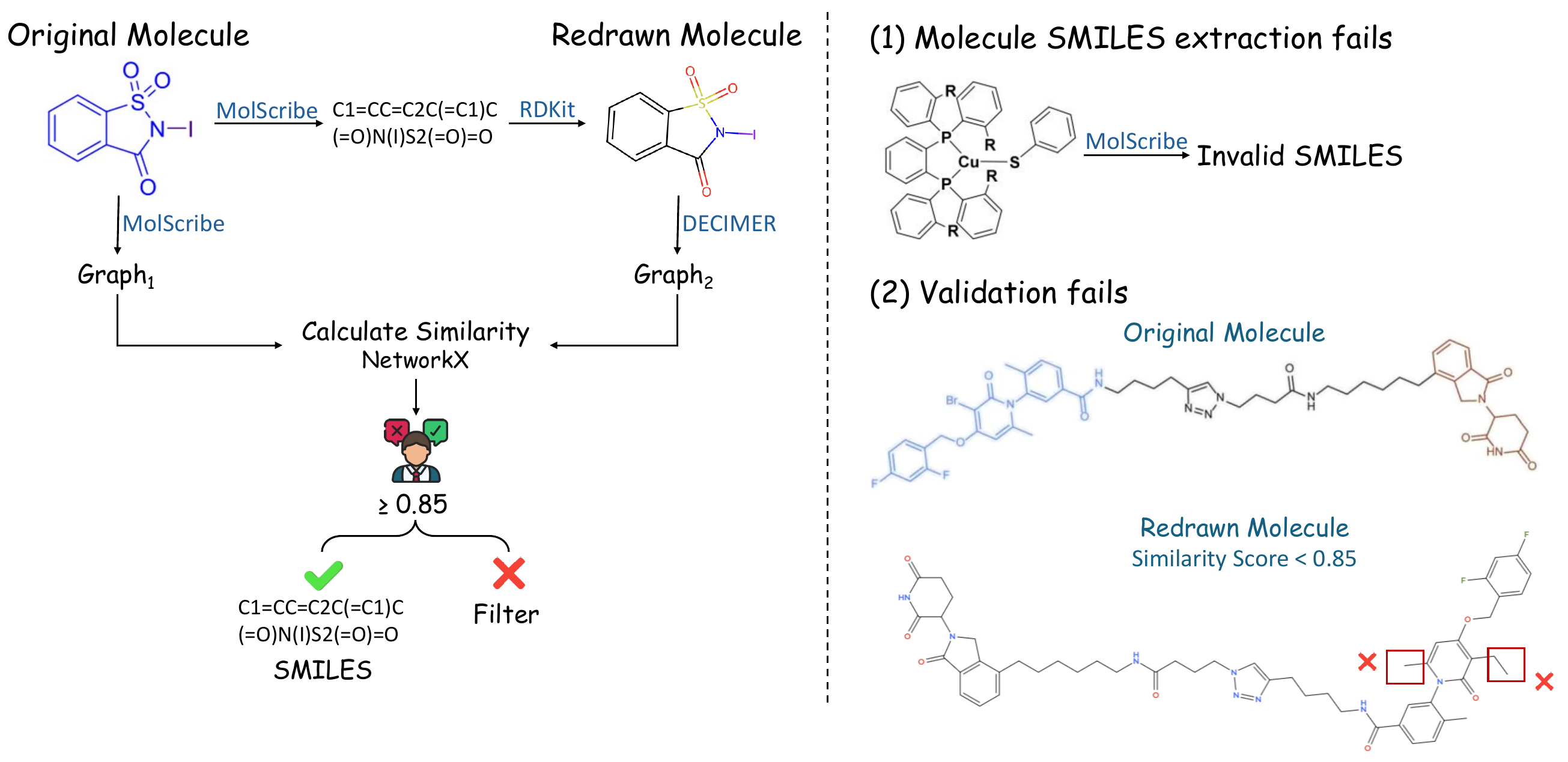}
  \caption{SMILES extraction and round-trip validation in the SMILES-first pipeline (left), with two representative failure modes (right): (1) MolScribe fails to extract a valid SMILES from a structurally complex molecule; (2) extraction succeeds but the regenerated structure differs from the original, failing the similarity check.}
  \label{fig:smiles_failures}
\end{figure}

\begin{figure}[h]
  \centering
  \includegraphics[width=\textwidth]{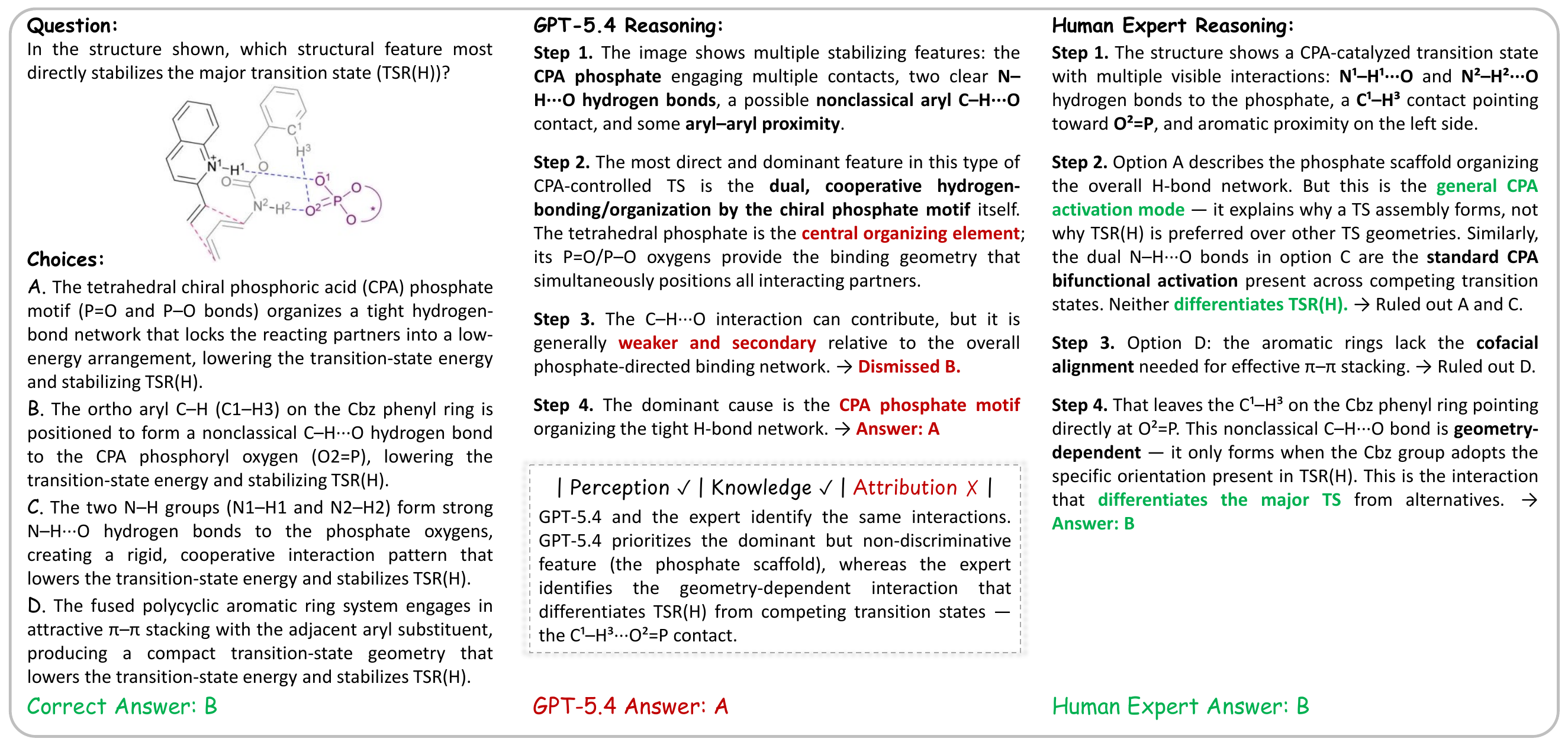}
  \caption{Representative case where GPT-5.4 commits an attribution error while Expert~A, answering without the source paper, correctly identifies the responsible structural feature within his domain.}
  \label{fig:case_human_wins}
\end{figure}

\begin{figure}[h]
  \centering
  \includegraphics[width=\textwidth]{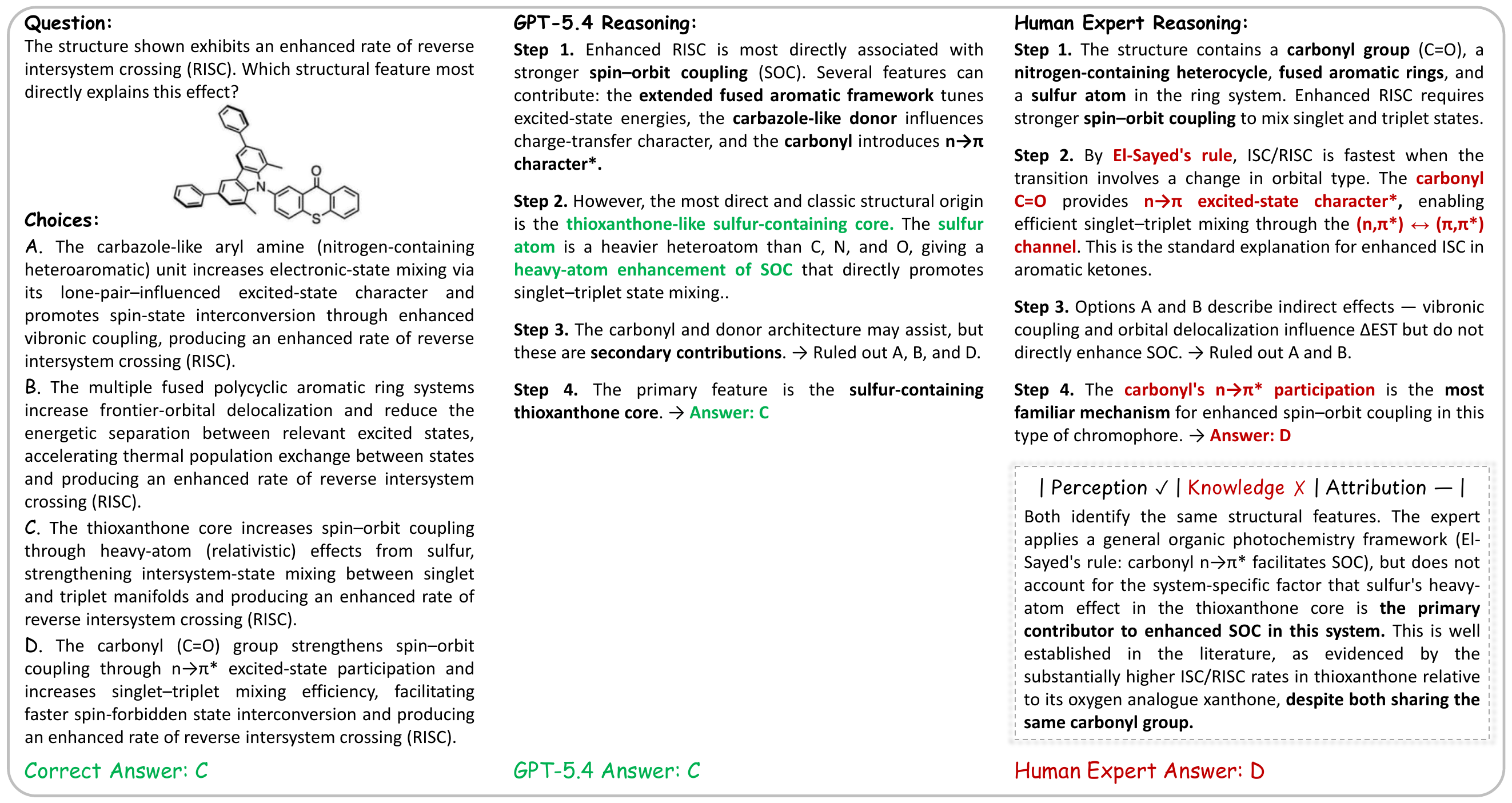}
  \caption{Representative case where GPT-5.4 answers correctly while Expert~A fails due to a domain knowledge gap on a question outside his primary expertise.}
  \label{fig:case_gpt_wins}
\end{figure}

\begin{figure}[h]
  \centering
  \includegraphics[width=0.95\textwidth]{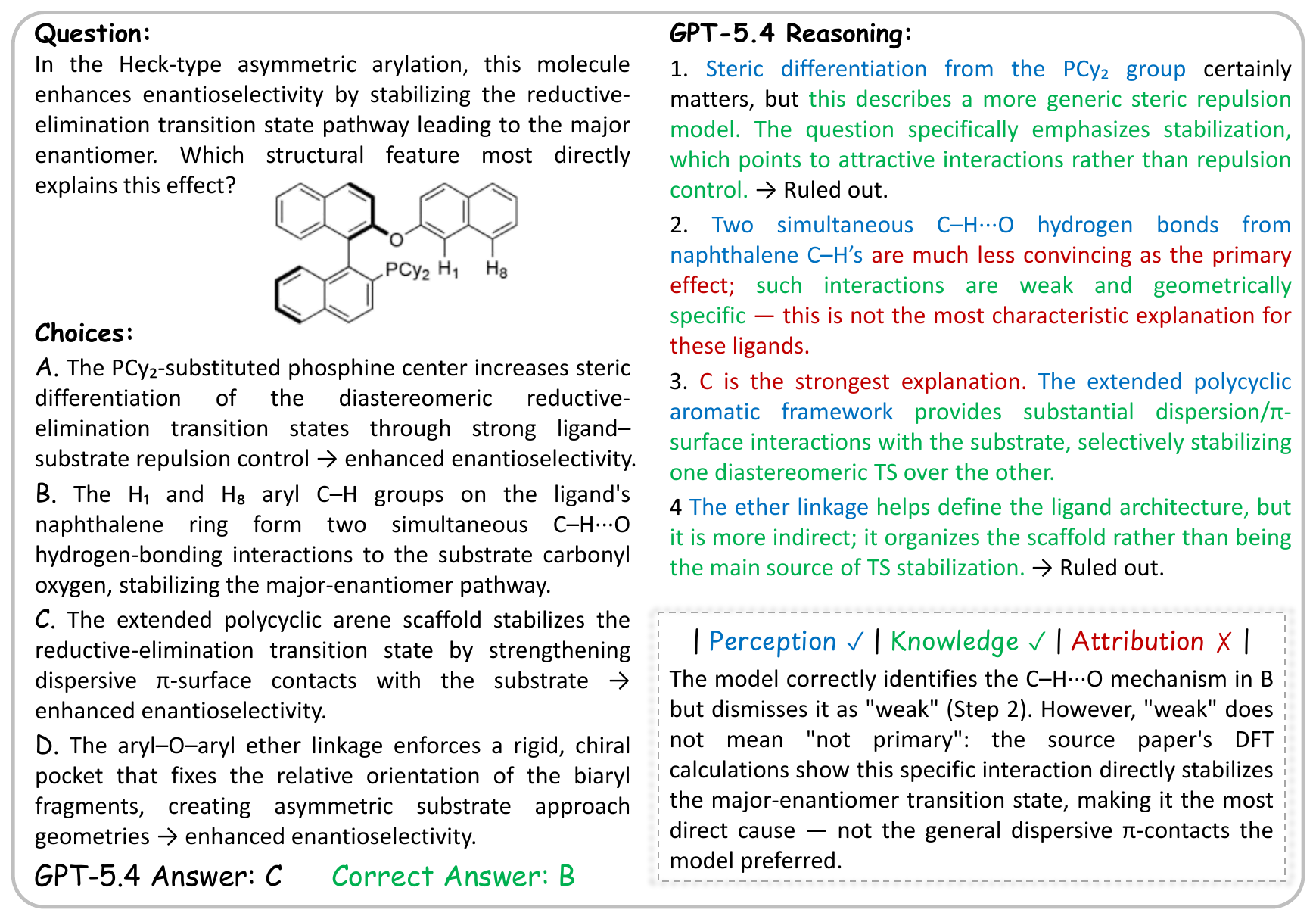}
  \caption{Representative attribution error from GPT-5.4 on \ourbenchmark{}. The model identifies the chemically relevant interaction and reasons plausibly about the options, yet selects the wrong feature as the primary explanation of the observed effect.}
  \label{fig:attribution_example}
\end{figure}

\begin{figure}[h]
  \centering
  \includegraphics[width=0.95\textwidth]{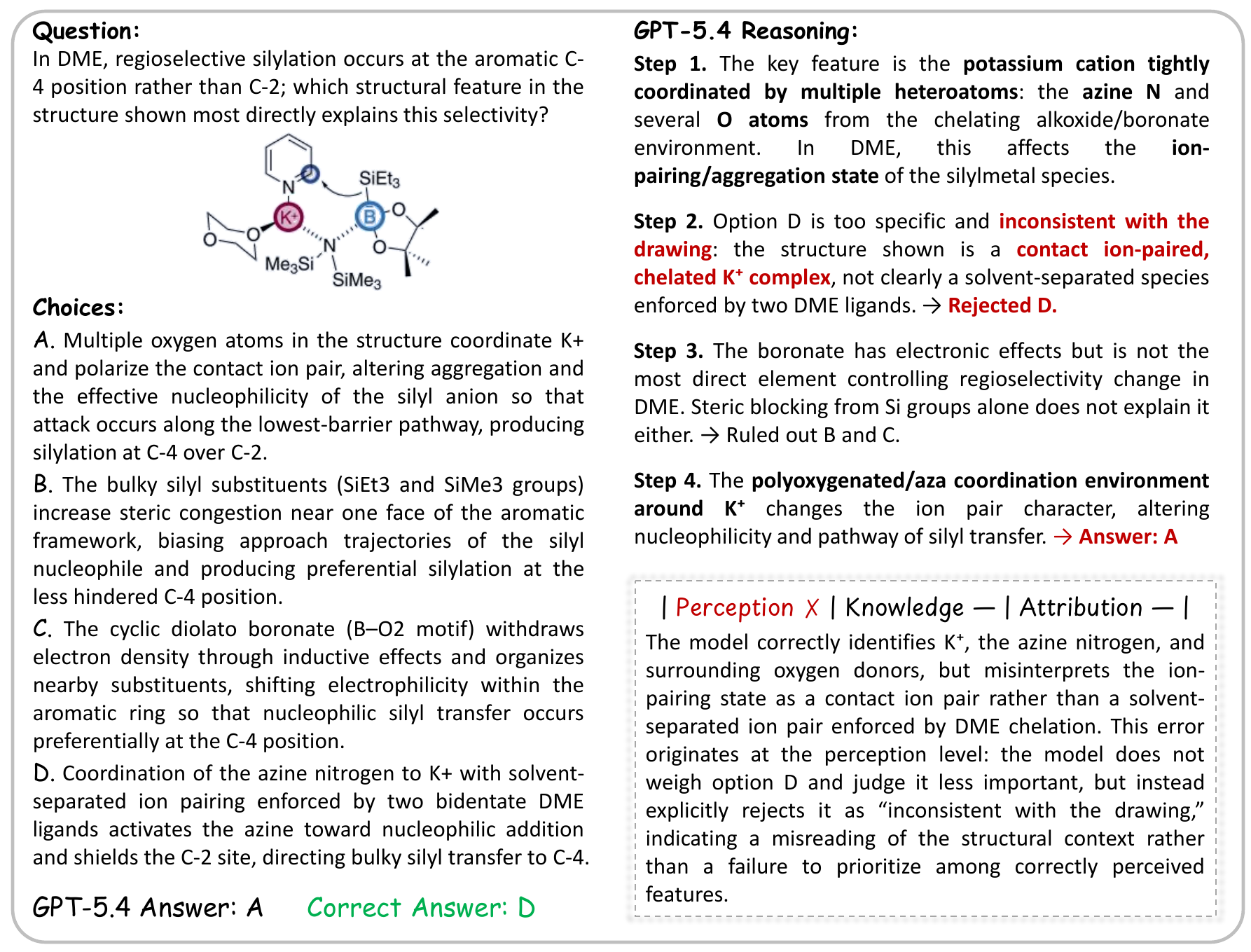}
  \caption{Perception error example. The model misidentifies or overlooks a structural feature visible in the image, leading to an incorrect answer.}
  \label{fig:case_perception}
\end{figure}

\begin{figure}[h]
  \centering
  \includegraphics[width=0.95\textwidth]{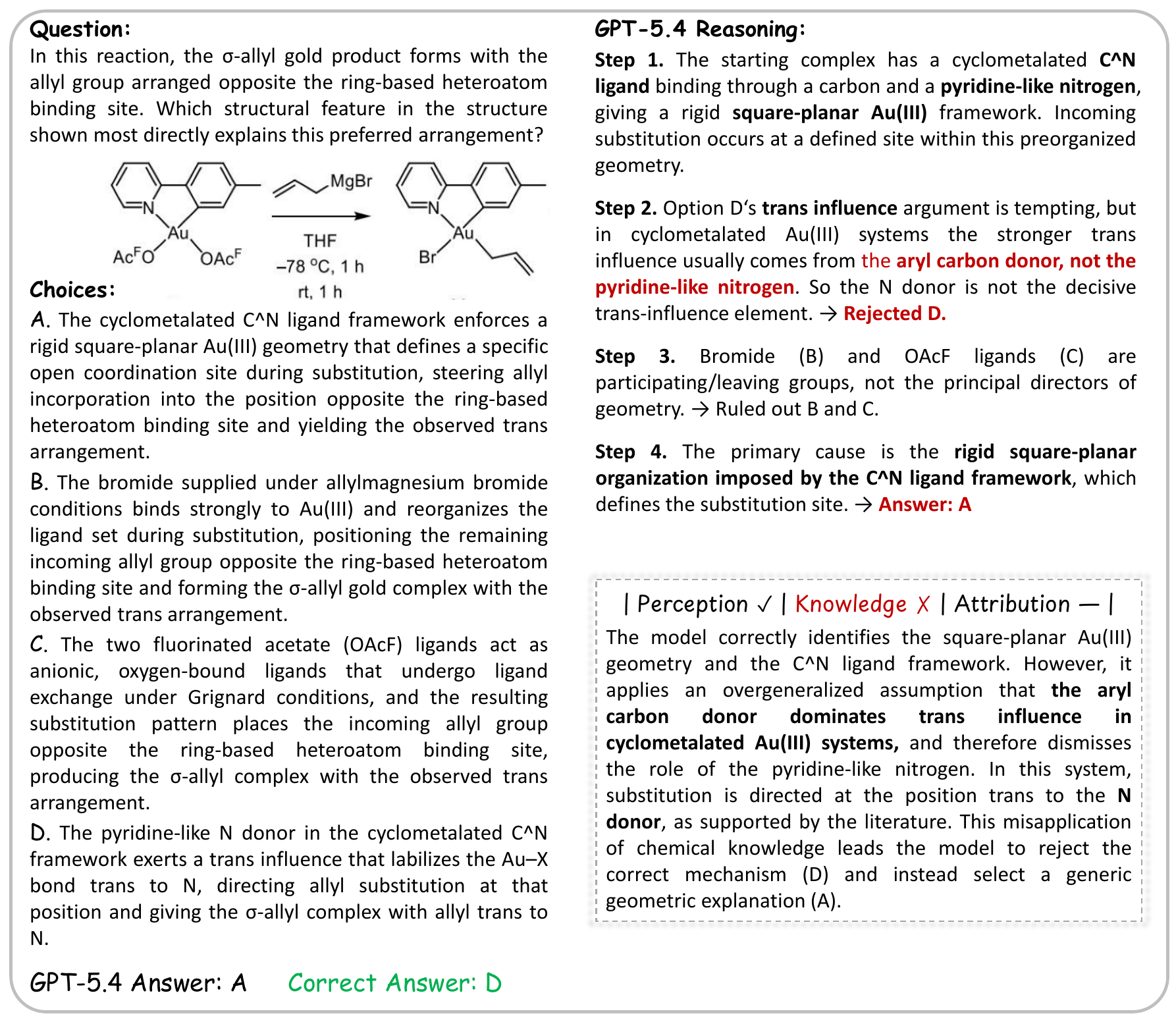}
  \caption{Knowledge gap example. The model correctly perceives the structural features but lacks the chemical knowledge to connect them to the observed effect.}
  \label{fig:case_knowledge}
\end{figure}

\begin{figure}[h]
  \centering
  \includegraphics[width=0.93\textwidth]{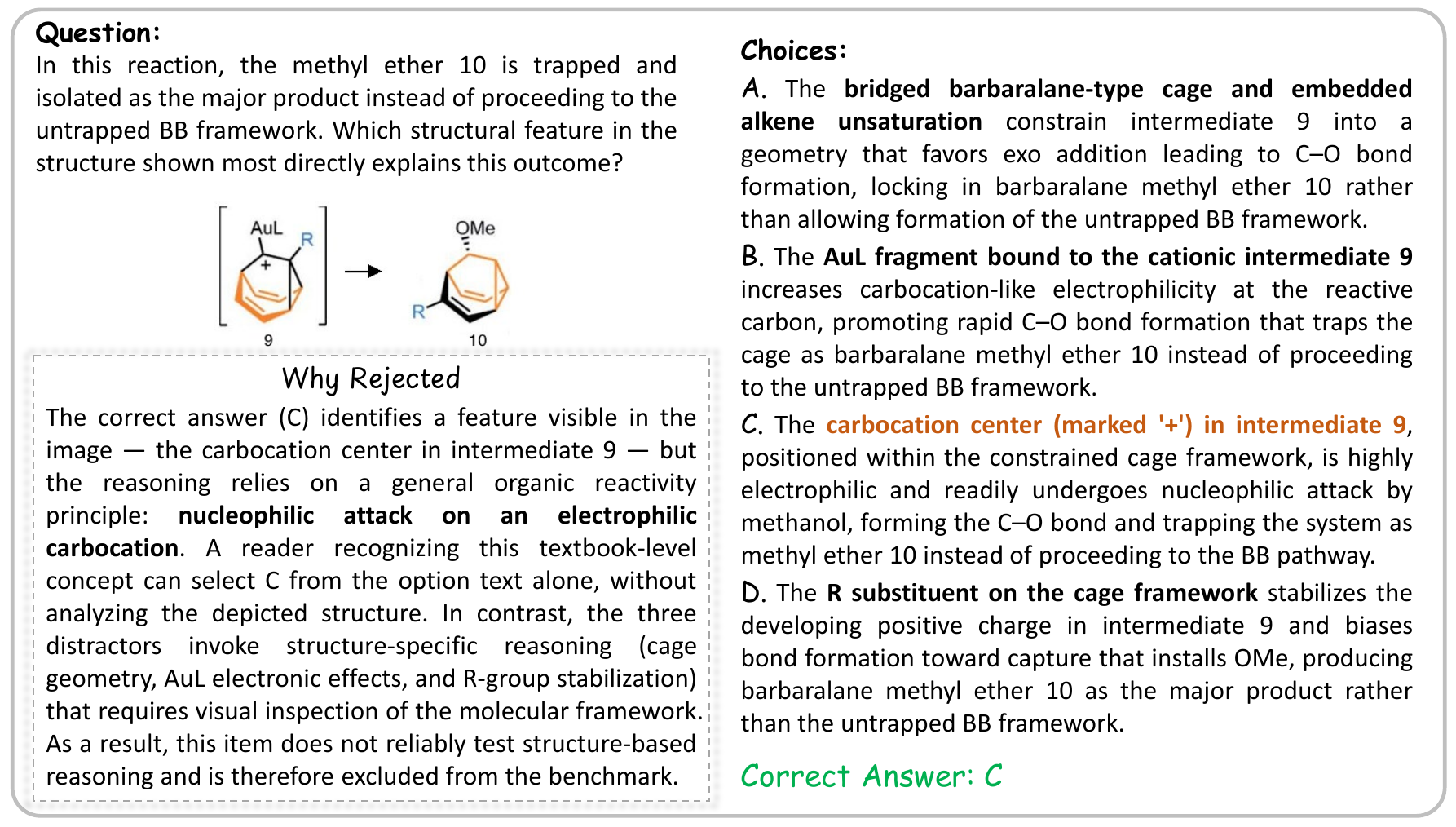}
  \caption{Structure necessity rejection (76.7\% of rejections). The correct answer can be selected without analyzing the depicted structure.}
  \label{fig:failure_necessity}
\end{figure}

\begin{figure}[h]
  \centering
  \includegraphics[width=0.93\textwidth]{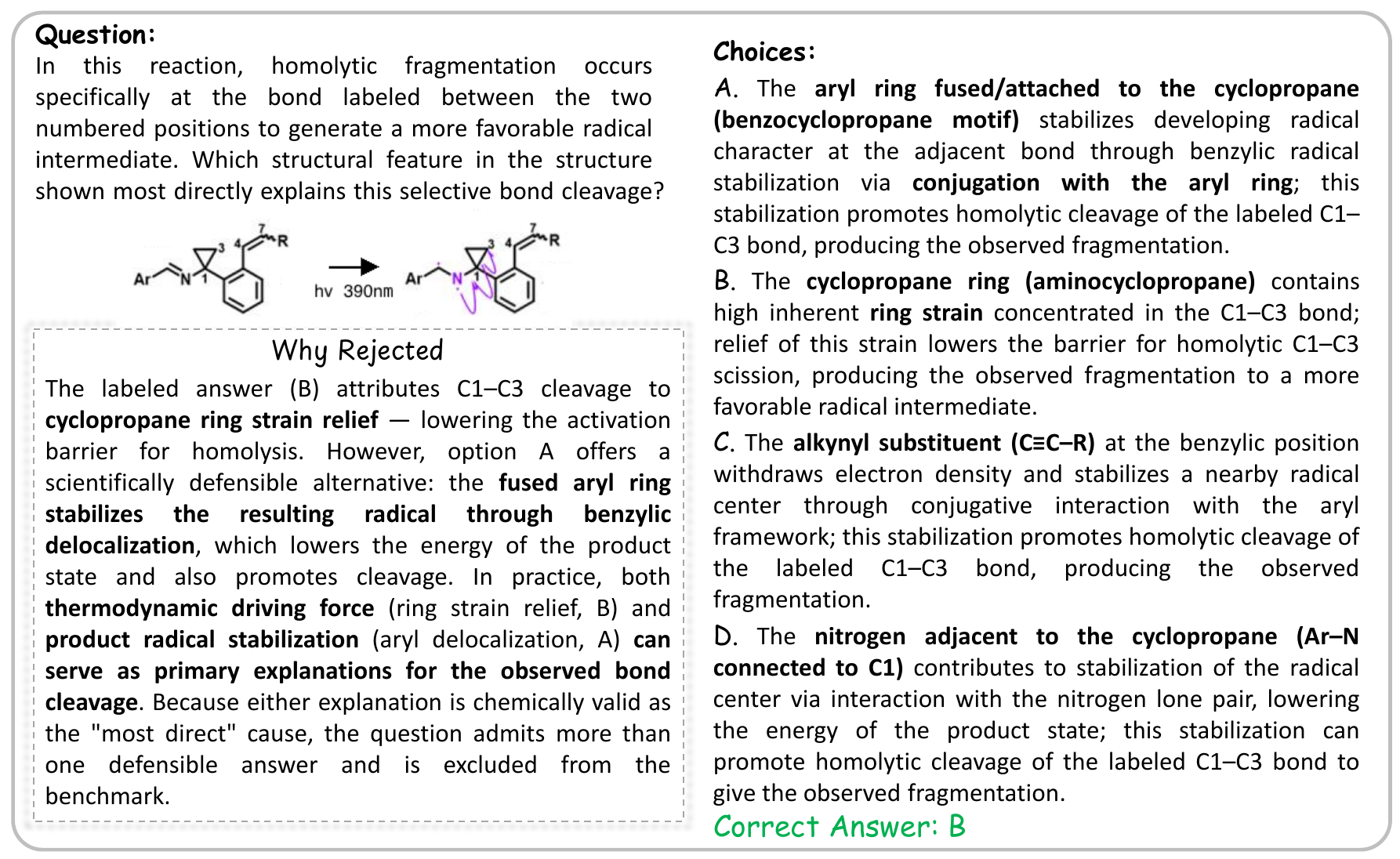}
  \caption{Answer unambiguity rejection (20.0\% of rejections). More than one option is scientifically defensible.}
  \label{fig:failure_ambiguity}
\end{figure}

\begin{figure}[h]
  \centering
  \includegraphics[width=\textwidth]{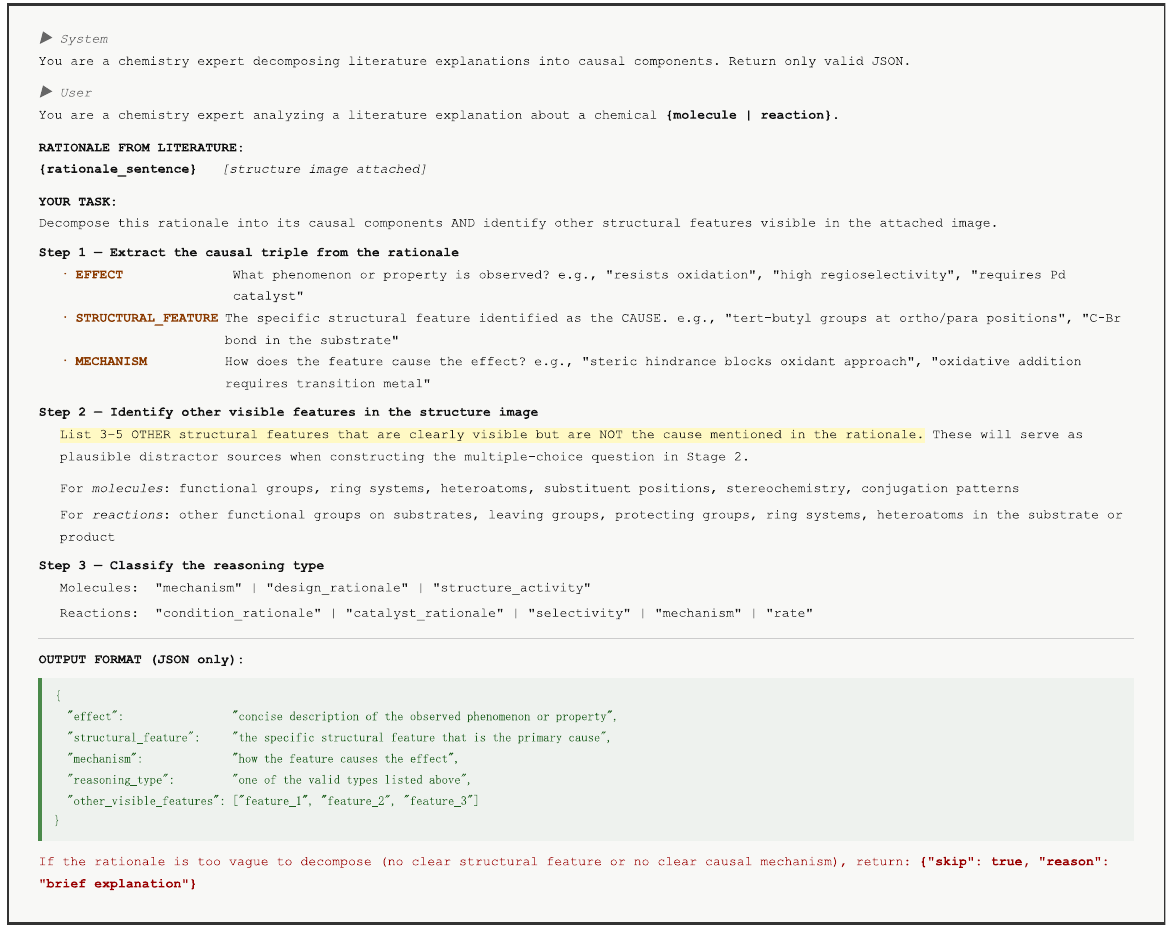}
  \caption{Prompt template for Stage 1 explanatory decomposition.}
  \label{fig:prompt_decomposition}
\end{figure}

\begin{figure}[h]
  \centering
  \includegraphics[width=\textwidth]{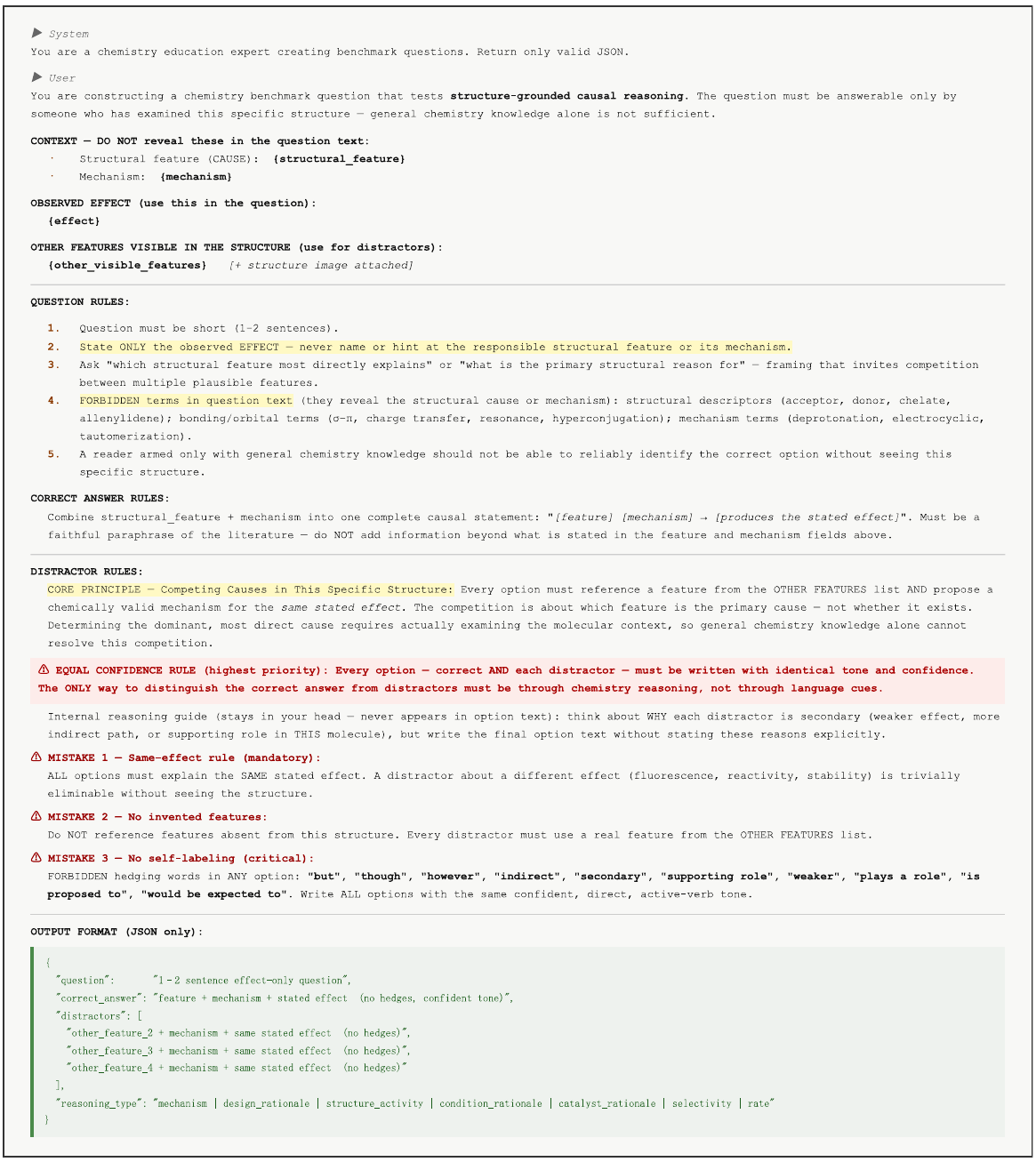}
  \caption{Prompt template for Stage 2 question and distractor generation.}
  \label{fig:prompt_generation}
\end{figure}

\begin{figure}[h]
  \centering
  \includegraphics[width=\textwidth]{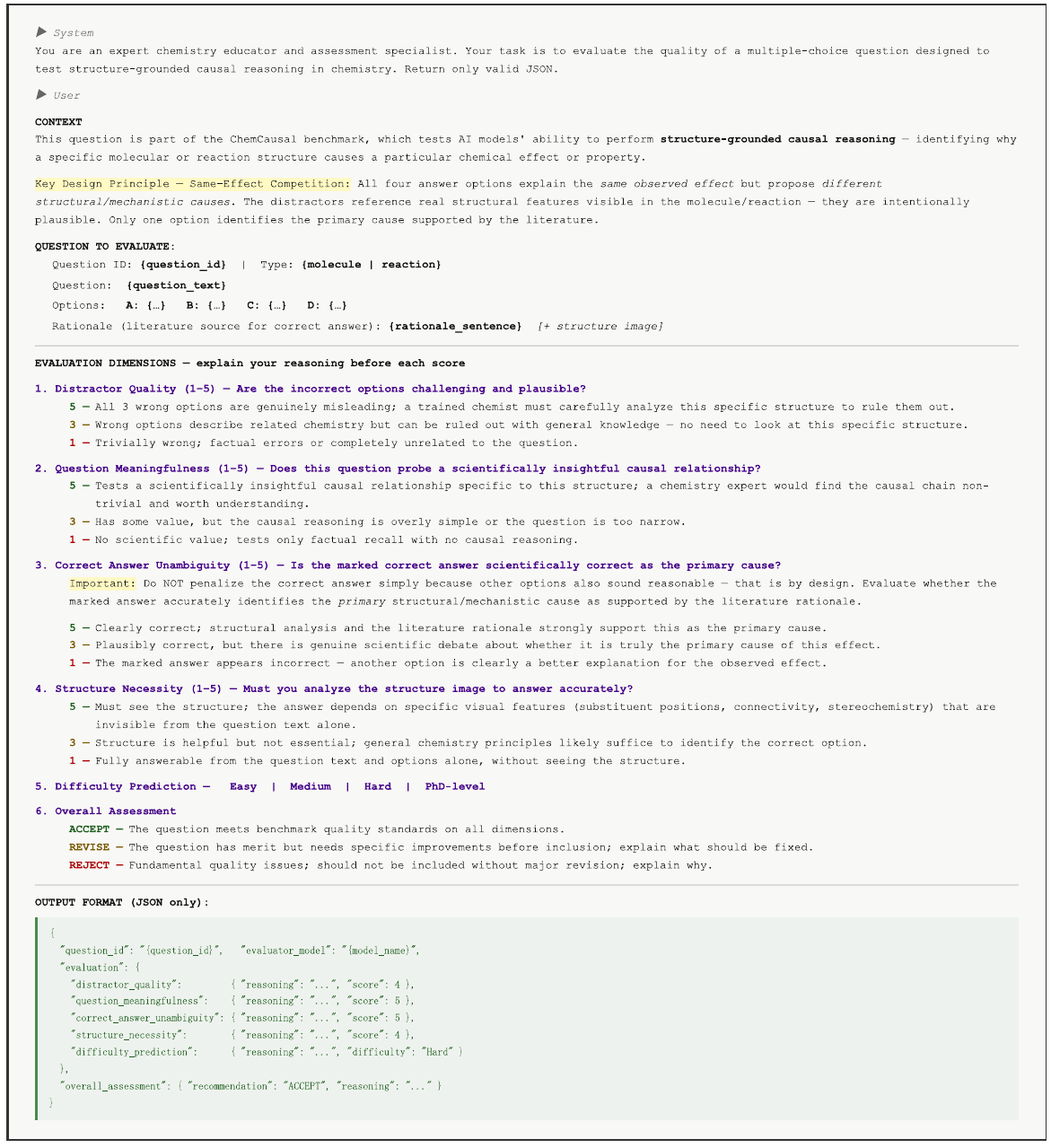}
  \caption{Prompt template for LLM-as-a-Judge quality screening on four dimensions.}
  \label{fig:prompt_judge}
\end{figure}

\begin{figure}[h]
  \centering
  \includegraphics[height=0.95\textheight]{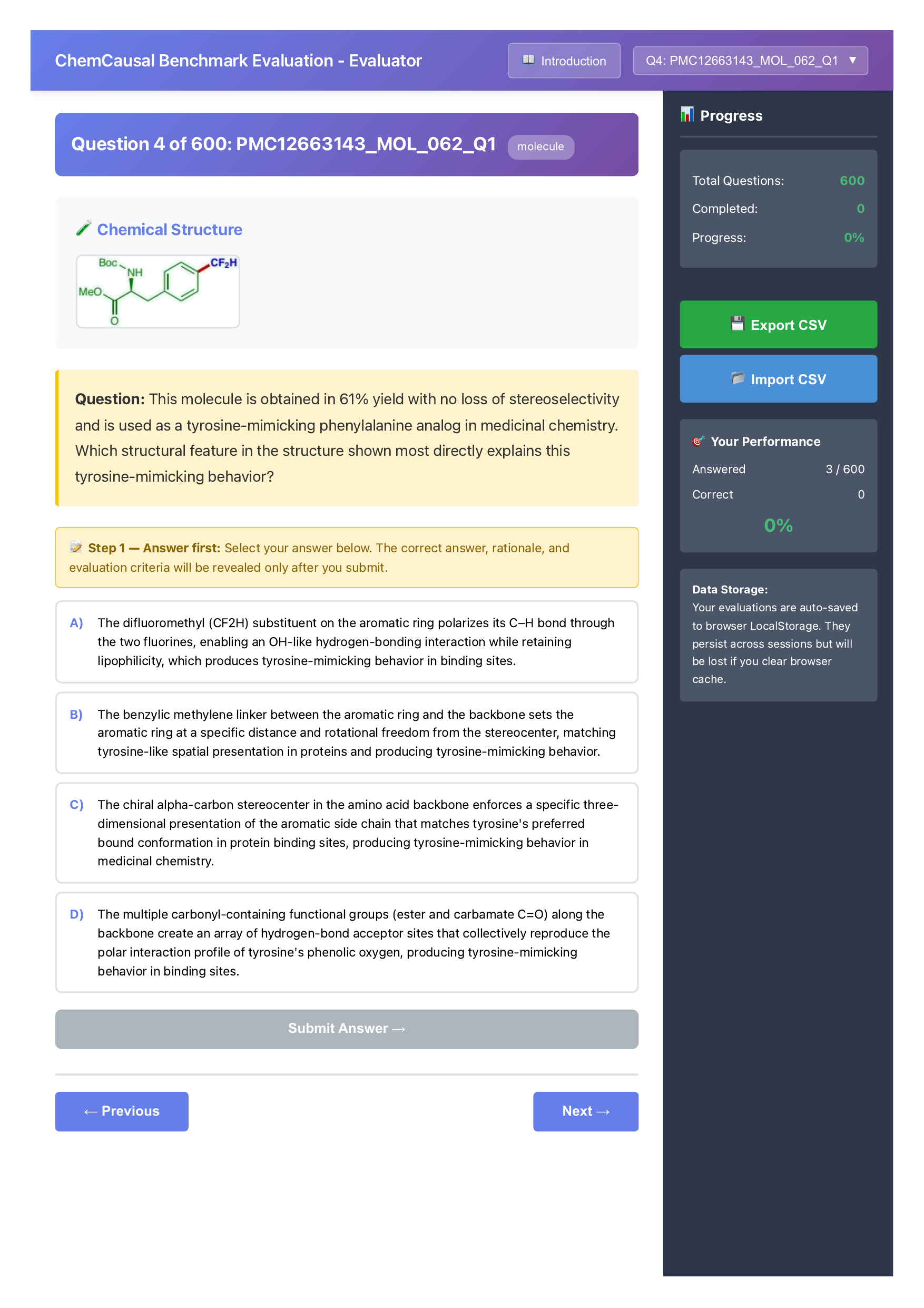}
  \caption{Human evaluation platform: blind answer phase, in which experts answer the question without access to the source paper text.}
  \label{fig:eval_platform_1}
\end{figure}

\begin{figure}[h]
  \centering
  \includegraphics[height=0.95\textheight]{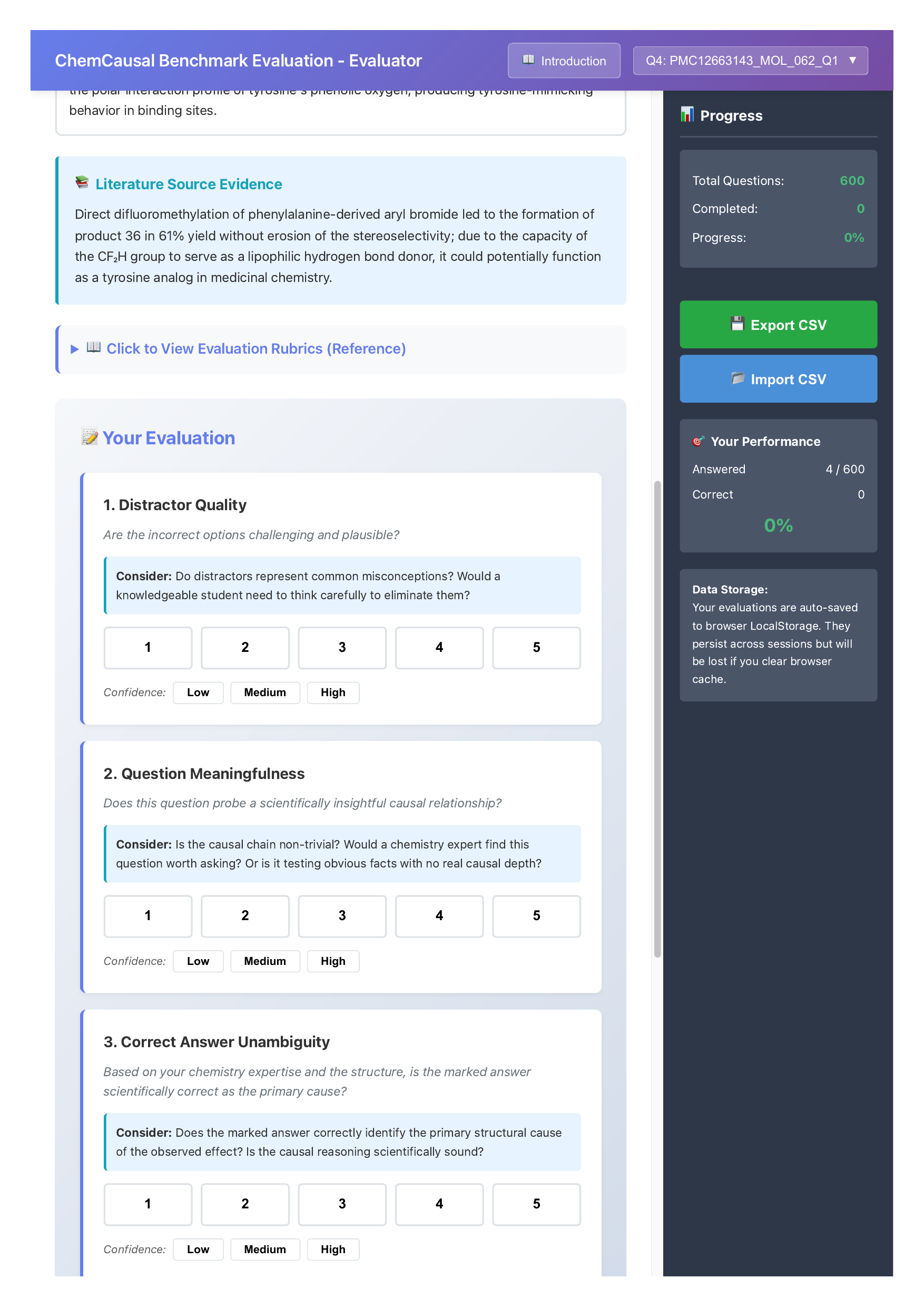}
  \caption{Human evaluation platform: evidence review and scoring phase, in which experts review the source paper evidence and score the four quality dimensions on a 1--5 Likert scale.}
  \label{fig:eval_platform_2}
\end{figure}

\begin{figure}[h]
  \centering
  \begin{minipage}{0.42\textwidth}
    \centering
    \includegraphics[width=\textwidth]{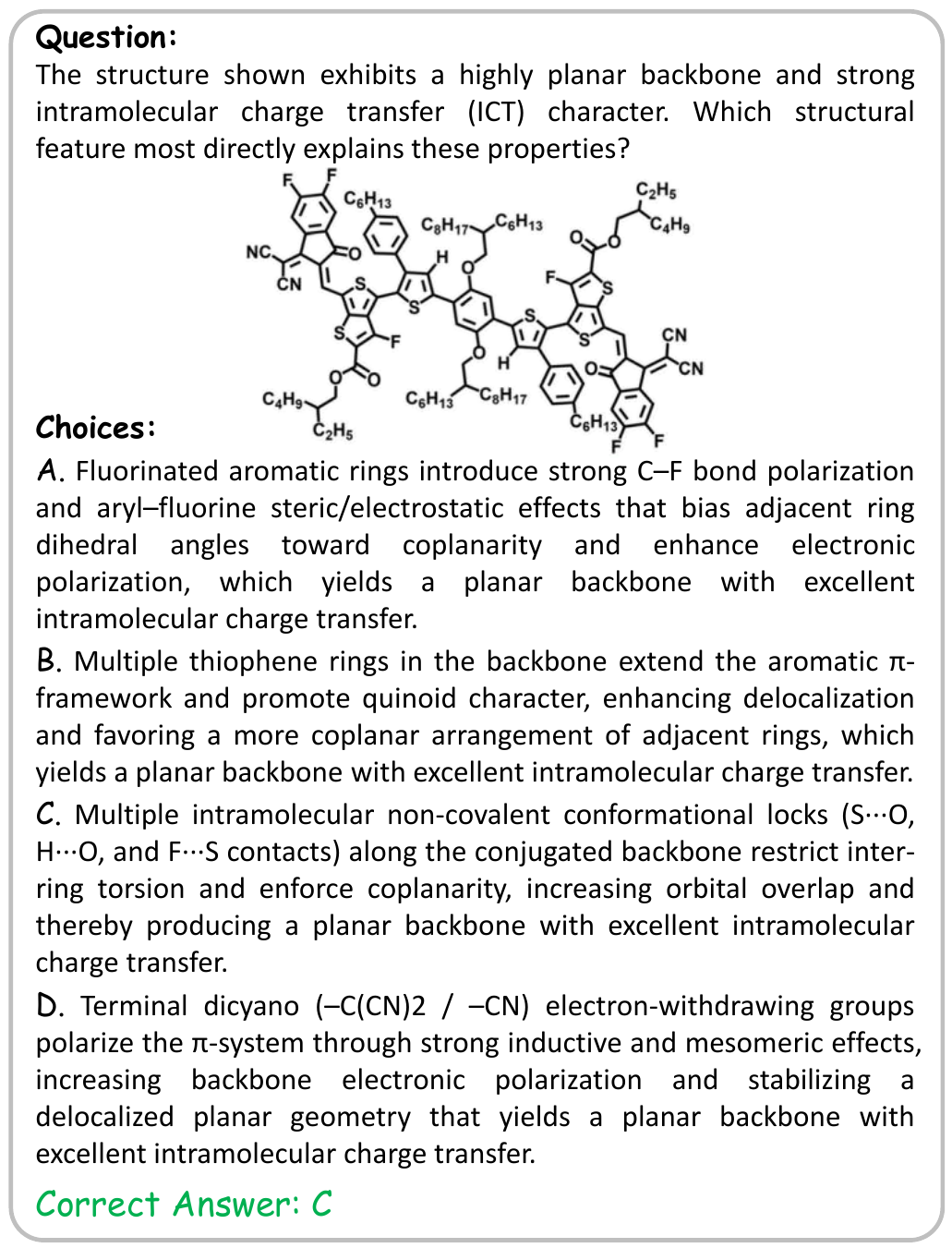}
    \caption{Mechanistic question on a molecule.}
    \label{fig:example_mech_mol}
  \end{minipage}
  \hfill
  \begin{minipage}{0.55\textwidth}
    \centering
    \includegraphics[width=\textwidth]{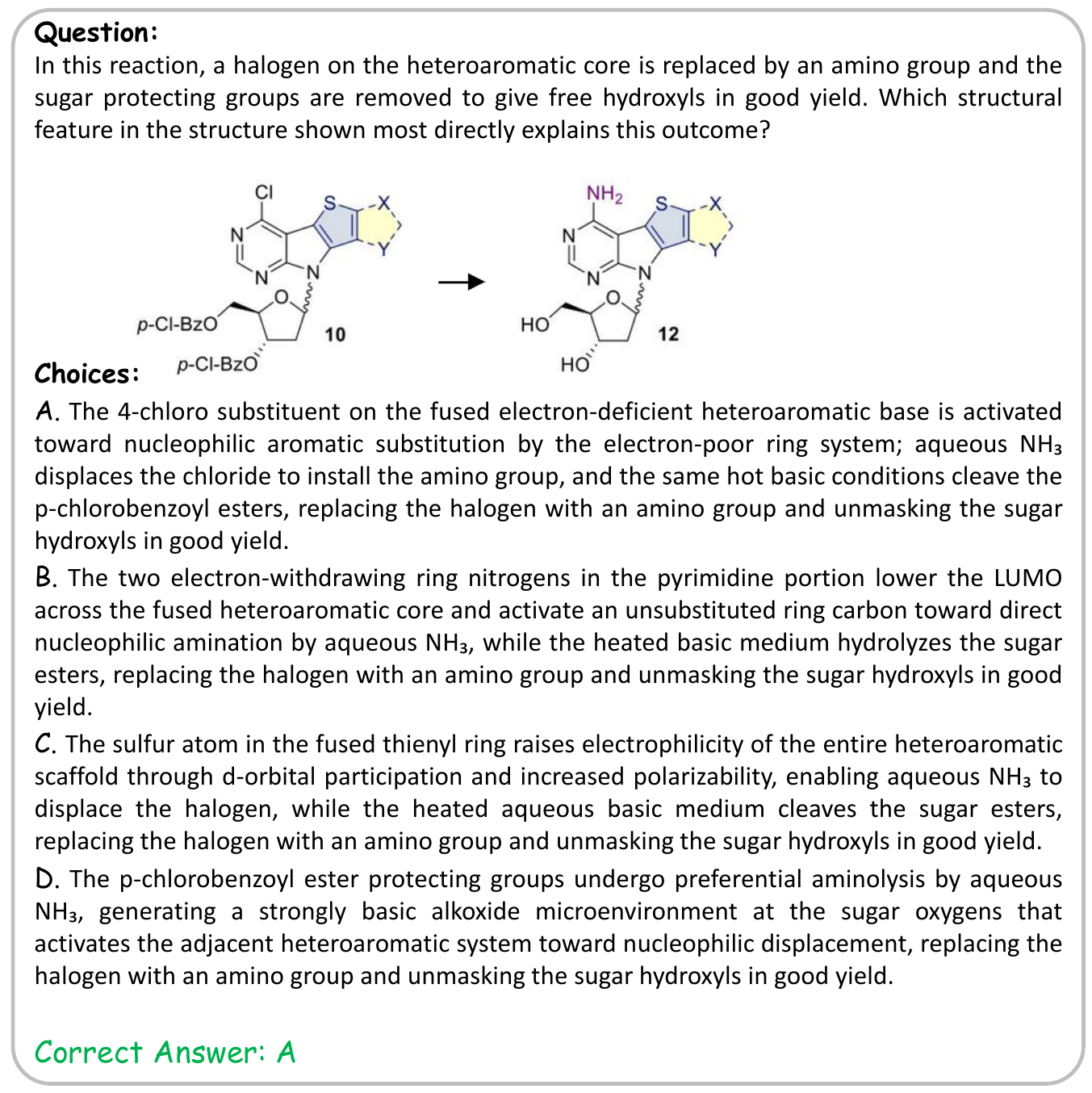}
    \caption{Mechanistic question on a reaction.}
    \label{fig:example_mech_rxn}
  \end{minipage}
\end{figure}

\begin{figure}[h]
  \centering
  \begin{minipage}{0.46\textwidth}
    \centering
    \includegraphics[width=\textwidth]{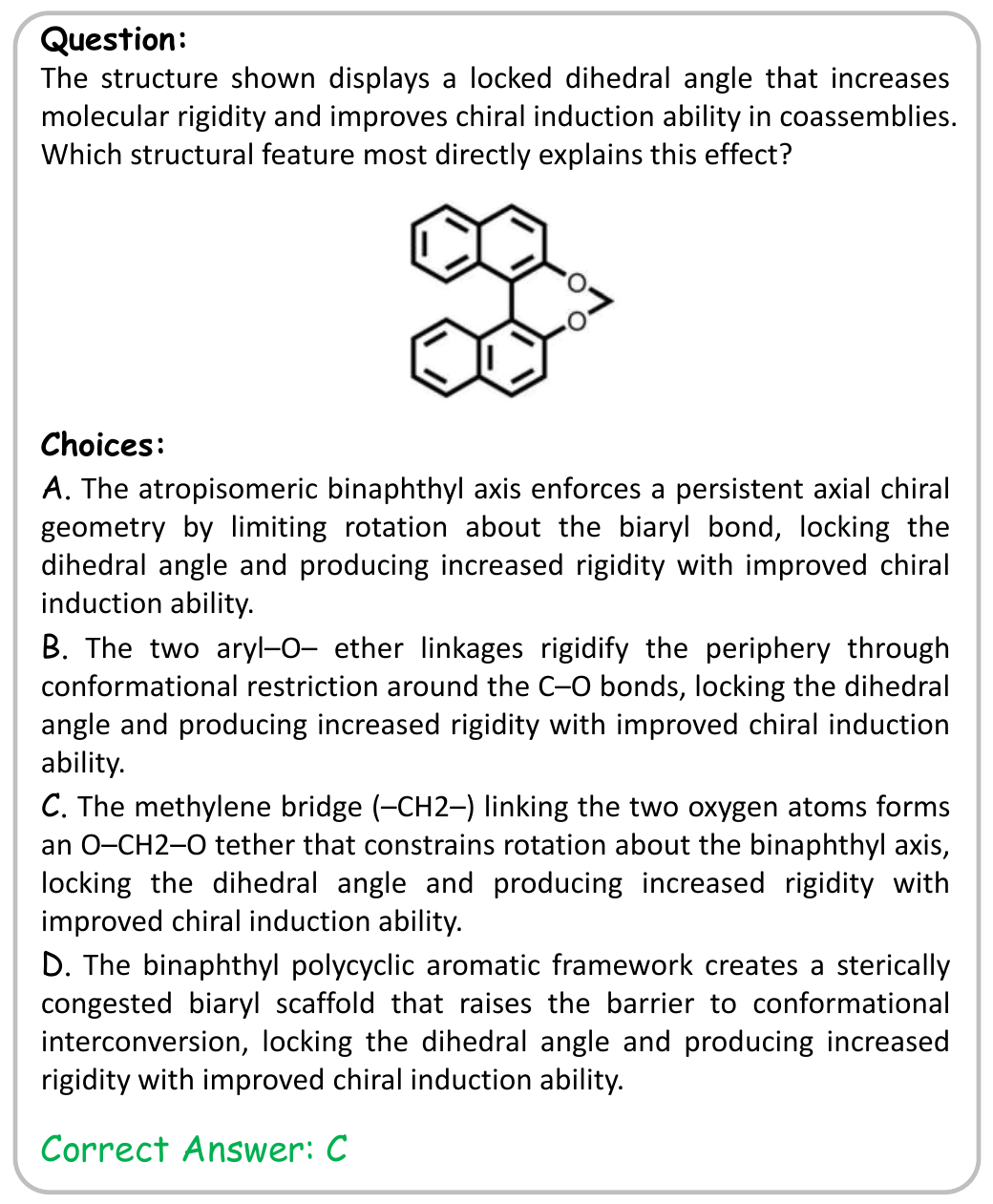}
    \caption{Design question on a molecule.}
    \label{fig:example_design_mol}
  \end{minipage}
  \hfill
  \begin{minipage}{0.5\textwidth}
    \centering
    \includegraphics[width=\textwidth]{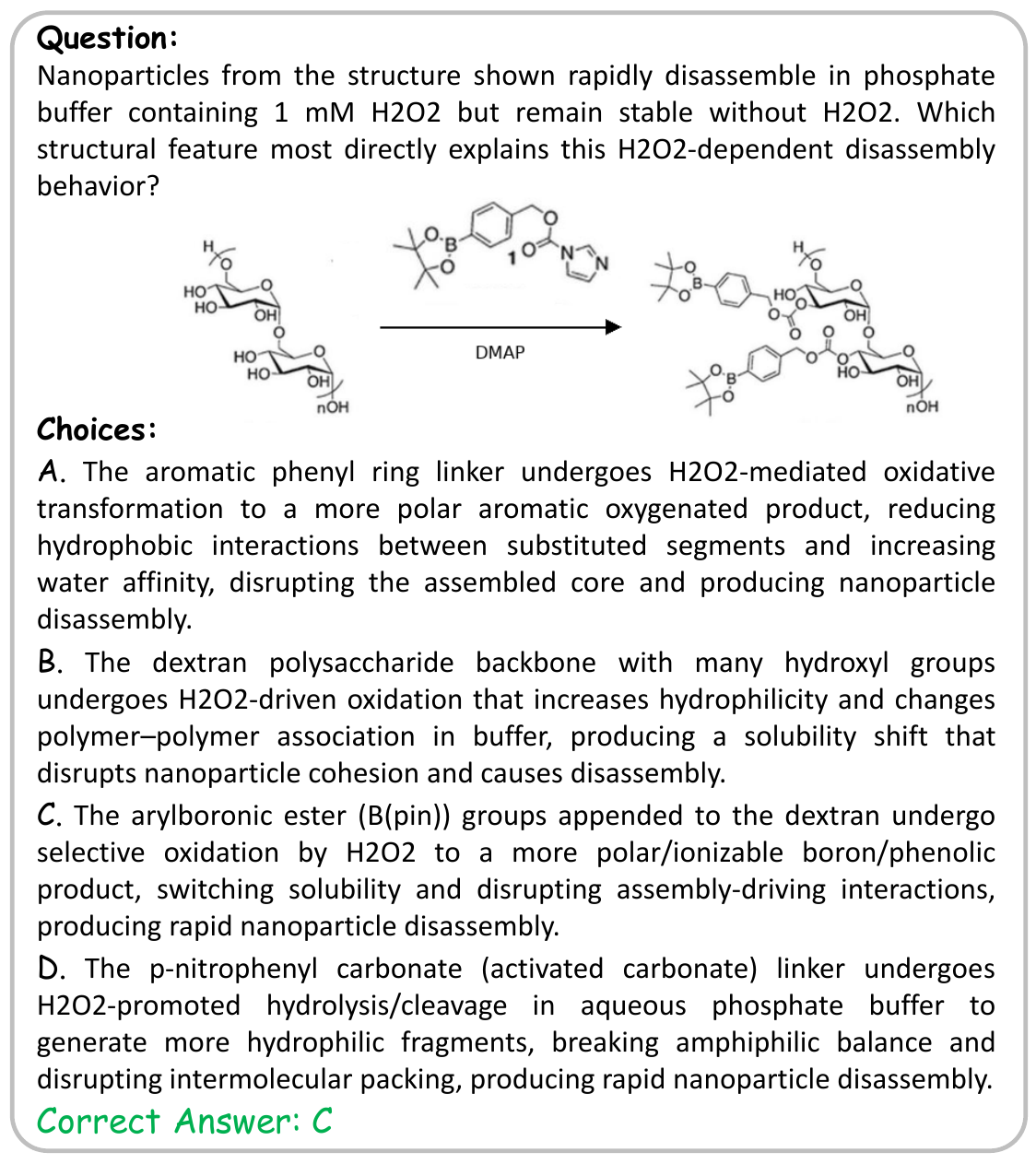}
    \caption{Design question on a reaction.}
    \label{fig:example_design_rxn}
  \end{minipage}
\end{figure}

\twocolumn

\begin{figure}[h]
  \centering
  \includegraphics[width=0.5\textwidth]{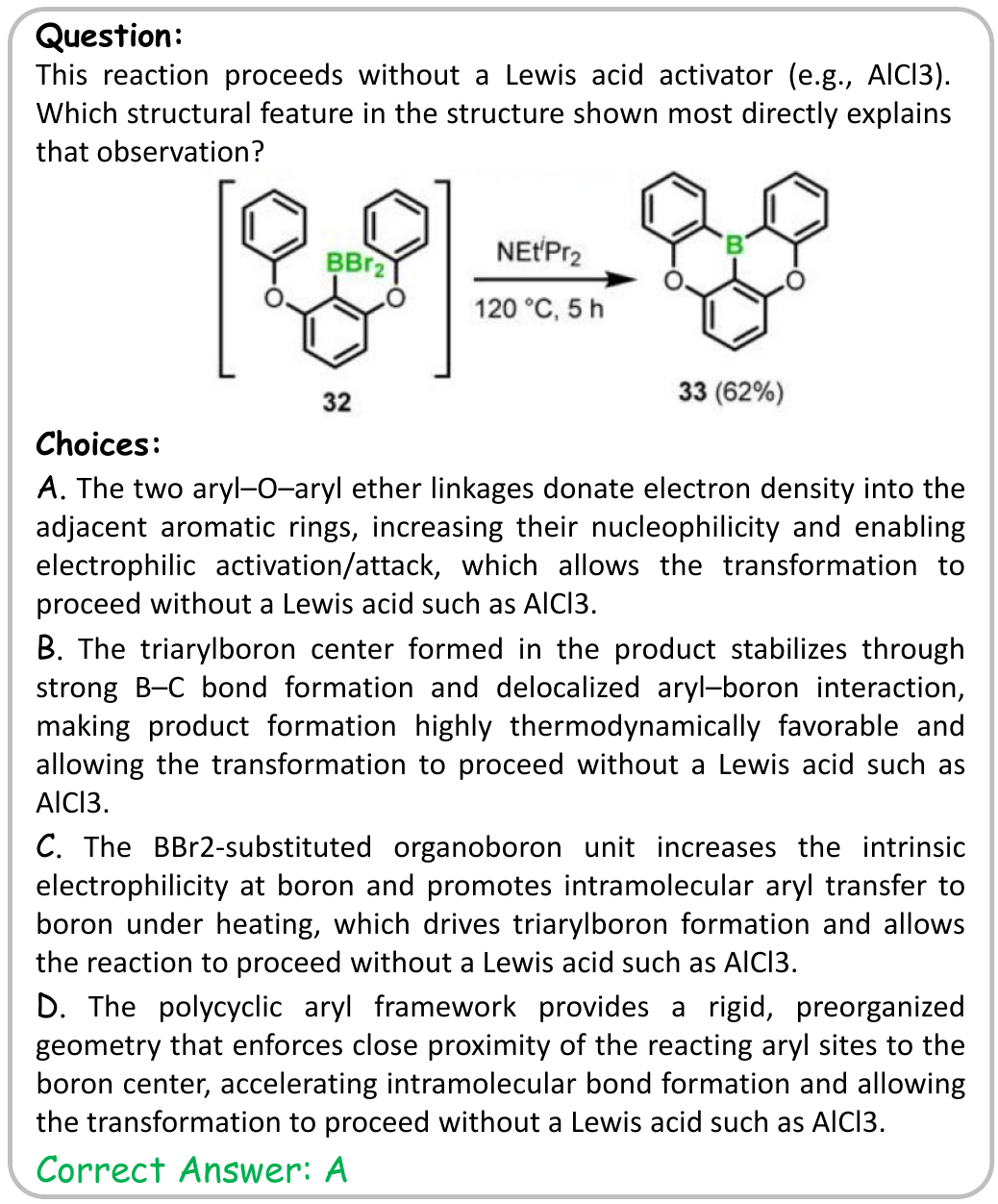}
  \caption{Condition question on a reaction.}
  \label{fig:example_cond_rxn}
\end{figure}

\end{document}